\documentclass[journal]{IEEEtran}
\ifCLASSINFOpdf
\else
\fi
\usepackage{graphicx}
\usepackage{amsmath,bm}
\usepackage{cleveref}
\usepackage{amsfonts}
\usepackage{setspace}
\usepackage{enumerate}
\usepackage{multirow}
\usepackage{booktabs}
\usepackage{color}
\usepackage{subfig}
\usepackage[square, comma, sort&compress, numbers]{natbib}
\usepackage{url}

\DeclareMathOperator*{\argmin}{\mathrm{argmin}}

\begin{document}

\title{How Does the Low-Rank Matrix Decomposition Help Internal and External Learnings for Super-Resolution}
\author{Shuang Wang,~\IEEEmembership{Member,~IEEE}, Bo Yue, Xuefeng Liang\IEEEauthorrefmark{1} and Licheng~Jiao,~\IEEEmembership{Senior~Member,~IEEE}
\thanks{Shuang Wang, Bo Yue and Licheng~Jiao are with Key Laboratory of Intelligent Perception and Image Understanding of Ministry of Education, International Research Center for Intelligent Perception and Computation, Xidian University, Xi'an, Shaanxi Province 710071, China.(e-mail:shwang.xd@gmail.com; yuebo312@live.com; jlcxidian@163.com)}
\thanks{Xuefeng Liang is with IST, Graduate School of Informatics, Kyoto University, Kyoto, 606-8501 Japan. e-mail: xliang@i.kyoto-u.ac.jp \IEEEauthorblockA{\IEEEauthorrefmark{1}Corresponding author}}} 


\maketitle
\begin{abstract}
Wisely utilizing the internal and external learning methods is a new challenge in super-resolution problem. To address this issue, we analyze the attributes of two methodologies and find two observations of their recovered details: 1) they are complementary in both feature space and image plane, 2) they distribute sparsely in the spatial space. These inspire us to propose a low-rank solution which effectively integrates two learning methods and then achieves a superior result. To fit this solution, the internal learning method and the external learning method are tailored to produce multiple preliminary results. Our theoretical analysis and experiment prove that the proposed low-rank solution does not require massive inputs to guarantee the performance, and thereby simplifying the design of two learning methods for the solution. Intensive experiments show the proposed solution improves the single learning method in both qualitative and quantitative assessments. Surprisingly, it shows more superior capability on noisy images and outperforms state-of-the-art methods.
\end{abstract}


\IEEEpeerreviewmaketitle

\section{Introduction}

\IEEEPARstart{S}{ingle-image} super-resolution (SISR) aims at reconstructing a high resolution (HR) image from a single low resolution (LR) image of the same scene. The ultimate goal is to obtain a visually pleasing image with high resolution, more details, and higher signal-to-noise ratio (SNR). It spans a wide bank of applications from visible image (satellite and aerial image, medical image, biometric image, etc.) to invisible image (ultrasound and range images, etc.). Unfortunately, SISR is ill-posed because it is an under-determined problem, thus
 may not result in an unique solution. To constrain the solution, the prior knowledge about the desired image is expected to be learned from the given data. To this end, SISR algorithms have stepped forward from interpolation to learning the mapping functions between the given LR image and the desired HR image. Recently, there exist two trends: \textit{external learning} and \textit{internal learning} for this purpose.

The external learning \cite{yang2008image,yang2010image,wang2012semi,cui2014deep,dong2014learning,zhang2015learning,dai2015jointly,dong2016accelerating,Kim_2016_VDSR} attempts to encode the expected information between aligned LR and HR image/patch pairs. Afterwards, the underlying scene can be obtained by decoding. Routinely, such information is learned from massive and varied external datasets, because these methods assume that the input LR image provides insufficient information. Among sophisticated machine learning algorithms, the coupled dictionary learning approaches reported a better performance. They are designed to extract the abundant ``meta-detail'' shared among external images which can be utilized for reconstructing details. This strategy brings us two pros: 1). As learned from external examples, the meta-detail consists of information which does not exist in the input LR image. Therefore, it is possible to enrich the lost details that are commonly shared in nature images; 2). If the external datasets are large scale, the meta-details have a rich diversity that is essential to ensure the quality of SR images. Unfortunately, the external learning cannot always guarantee a well recovery for an arbitrary input LR image. Especially, when certain singular and unique details rarely exist in the given datasets, external learning is apt to introduce either unexpected details or smoothness. This is called \textit{weak relevancy}. 

To avoid this issue, another attempt achieves SR images by exploring ``self-similarities'' among patches in the image \cite{protter2009generalizing,freedman2011image,giachetti2011real,huang2015single}. These methods, namely internal learning, stand on the fact that patches of a natural image recur within and across scales of the same image, and light a new way of image super-resolution. Consequently, the encoding of internal learning is built in the multi-scale and geometric invariant space between LR and HR patch pairs.
When compared to the external learning, the internal learning may provide less quantity and diversity of feasible pairs, but they are more \textit{relevant} to the input LR image. Thus, it inherently outperforms external learning on the images/regions that contain the densely repeated patterns. As such, the core of internal learning is locating the self-similarities regardless of the fact that there are scale change and geometric distortion. However, inappropriate matches of irregular patterns in the image lead to artifacts. 

Obviously, neither the external learning nor the internal learning is perfect for SISR. We question whether an integration of the two could enhance SISR performance? By analyzing their results in the feature space (\textit{estimation error distribution}) and the image plane (\textit{preference map}), our finding reveals that their attributes are quite different. Specifically, two methods produce rather complementary details in the feature space and the image plane (see Section III A for details). However, the straightforward combination, e.g. a weighted average, may not reach an improvement because of the unknown balance of priors and errors from two learning methods. Wang et al. \cite{wang2015learning} proposed an alternative method to fuse an internal HR result and another external HR result according to the preference by comparing their SNRs on each small patch. However, there still exist two remaining problems. 1). \textit{Less diversity}: the method fuses only two inputs which makes the result heavily relies on the chosen methods and datasets. 2). \textit{Biased criterion}: SNR is just one of many criteria in super-resolution problem. It may not always result in a visually pleasing outcome.

\begin{figure*}[!t]
\setlength{\abovecaptionskip}{-0cm}
\setlength{\belowcaptionskip}{-0.5cm}
\centering
	\includegraphics[width=1\linewidth]{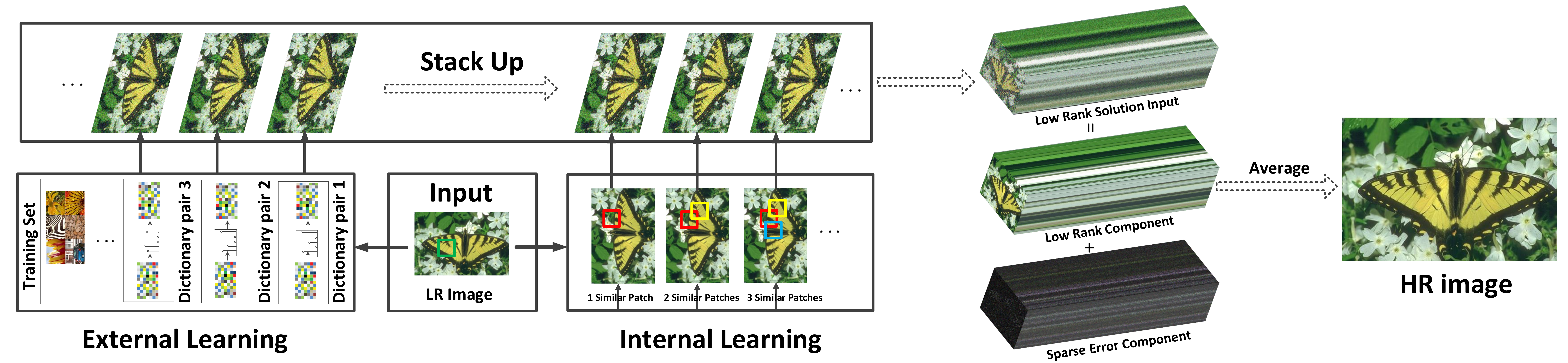}
\caption{The pipeline of the proposed low-rank solution.}
\label{fig:pipeline}
\end{figure*}

To address all above problems, we first increase the diversity. Specifically, an external learning method is tailored by learning the mapping function from varied datasets, which brings multiple preliminary HR images for improving the reliability. For internal learning, we tailor the patch matching criteria which gives weight to the similarity matching and consider the geometry transformation as well. It also produces multiple preliminary HR images. With dozens of HR images from both methods, we change the viewpoint to treat the data fusion as a dimension reduction problem and build these preliminary HR images into a matrix. Our analyses reveal the internal and external learning errors in the matrix are sparse and mostly full rank. This finding inspires us to employ the low-rank matrix decomposition for the integration, which can efficiently recover a low-rank matrix corrupted by errors and noises, while at the same time preserving the widespread pixels into the reconstructed image. Our method essentially differs from other combinations. It neither requires specific parameters nor relies on a single criterion, but fuses images according to their inherent properties to achieve a better qualitative and quantitative result. Thus, our contribution is threefold:

\begin{enumerate}
  \item We propose a parameter-free solution that considers the integration of internal and external learning for SISR as a low-rank matrix decomposition problem. We also tailor the external learning and internal learning methodologies to improve the diversity and fit to our solution.
  \item We theoretically and experimentally prove that only a few internal and external learning outputs can result in a desired SR image. It simplifies the design of the internal and external learning methods for the low-rank solution.
  \item Due to the inherent anti-noise attribute, our low-rank solution is robust on a wide variety of data, including noiseless images, synthetic noisy images and real noisy images.
\end{enumerate}

The experiment on the noiseless images shows our proposed low-rank solution outperforms the state-of-the-art methods on the benchmark databases. For the noisy images, our solution again demonstrates a superior ability of restraining noise. Even if the noise variance is up to 20.

In the remainder of the paper, we first briefly review the most relevant works and the state-of-the-art methods in section \ref{sec:2}, then, explore the attributes of internal and external learnings which are the foundation for the integration using low-rank matrix decomposition in section \ref{sec:3}. Section \ref{sec:4} gives the proposed low-rank solution in detail, and provides a theoretical analysis of its performance against input. Furthermore, we introduce how to tailor an internal learning method and an external learning method for the solution. In section \ref{sec:5}, intensive experiments show the effectiveness of the proposed solution on a variety of data and settings. We finally conclude in section \ref{sec:6}.

\section{Related Works}
\label{sec:2}
In SR literature, recent studies more focused on external learning methods  that learn priors from a huge volume of LR-HR image pairs to predict the HR image. For example, the seminal work, Freeman et al. \cite{freeman2002example} employed Markov Random Field to select the appropriate HR image patches from candidates. This approach is capable of producing plausible fine details. However, lack of relevant images in the database results in a fairly poor outcome. Yang et al. \cite{yang2010image,yang2012coupled} introduced a dictionary learning strategy for SR using sparse representation. They also assumed that the sparse representation of HR image patches are  similar to that of the LR one. These methods desire to learn two coupled dictionaries, namely, the over-complete HR dictionary and the LR dictionary. Being combined with the HR dictionary, HR image patches are able to be well constructed. Zeyde's method \cite{zeyde2012single} is based on Yang's framework, but reduces the computation cost. It represents the LR image patches by performing dimension reduction on features using PCA, and the HR dictionary is directly obtained by the pseudo-inverse. Timofte et al. \cite{timofte2013anchored,timofte2014a+} combined the anchored neighborhood regression with sparse representation for fast single image SR. Recently, deep networks (Deep Network Cascade \cite{cui2014deep} and Convolutions Neural Network (CNN) \cite{dong2014learning} are utilized to directly learn the end-to-end mapping between the LR/HR image patch pairs. Furthermore, Dong et al. \cite{dong2016accelerating} proposed a compact hourglass-shape CNN structure to further accelerate the SR, and Kim et al. [9] designed a very deep CNN for SR which resulted in a significant quantitative improvement.

On the contrary, internal learning methods achieve SR by exploring the local self-similarities in the image instead of using external data. Freedman et al. \cite{freedman2011image} super resolved a LR image using priors from the similar patches found in a neighbor region under the same scale. The quantity of matched patches, however, is not always sufficient for a desired SR result. Thus, Glasner et al. \cite{glasner2009super} introduced an approach of searching across varied image scales to increase the matching number. Recent studies reported above ¡°translated¡± matching is unlikely to handle the textural appearance variation. Therefore, Zhu et al. \cite{zhu2014single} deformed the patched using optical flow to allow texture change. Huang et al. \cite{zhu2014single} extended the internal patch search space by applying affine transformations for geometric variations in image patch.

In reality, the LR images usually suffer from noise corruption. However, until now, only a few researches focused on noisy image SR. Sigh et al. \cite{singh2014super} tried to integrate the merits of image denoising and image SR, and then proposed a convex combination of the denoised result and the SR result of a noisy input. Liu et al. \cite{liu2016robust} took advantage of the task transferring of deep learning. They first trained a SR network based on noiseless data, and subsequently fine-tuned the net by using the noisy data. As a result, their model was adapted to both noiseless and noisy tasks.

So far, works aimed at combining pros of both internal learning and external learning are still rare. The method \cite{wang2015learning} fuses an internal HR result and another external HR result according to the internal and external learning preference by comparing their SNRs on each small patch. It reported an improved performance on ultra high definition image. Recently, low-rank decomposition method has been applied in a wide variety of computer vision tasks, including image restoration \cite{dong2013nonlocal,zhang2014hyperspectral}, person re-identification \cite{jing2017super}, face recognition \cite{chen2012low, jing2016multi}, image classification \cite{zhang2011image, zhang2013learning, wu2016multi}, etc. To the best of our knowledge, this is the first work in the literature to use low-rank decomposition for integrating the internal learning and external learning methods in the image SR problem.
\section{Attributes of Internal Learning and External Learning}
\label{sec:3}
As aforementioned, the internal learning and external learning contribute differently to a SR result. It is certainly worth analyzing their own attributes, and exploring an appropriate methodology to integrate their pros. To this end, we follow a general model of SR problem but address on the complementarity and sparsity of two learning methods.

In general, SR problem can be formulated as
\begin{equation}
\hat{X}= \argmin \Vert X-f(Y)\Vert_p^p + \lambda R=X + E,
\label{eq1}
\end{equation}
where $\hat{X}$ is the super resolved image, $X$ is the ground truth HR image, $Y$ is the given LR image, $f$ is the SR function, $1\leq p\leq 2$, $\lambda$ is the balance weight, $R$ is the regularization in SR function, and $E$ is the estimation error map with the same size of $X$. Although the recovered detail is implicit in Eq. (\ref{eq1}), it is correlated with $E$ because $X$ and $Y$ are known. Thus, studying the attribute of a learning method is equal to studying $E$.

\subsection{Complementarity Analysis}
\label{sec:3.1}
\begin{figure}[!t]
\setlength{\abovecaptionskip}{-0cm}
\setlength{\belowcaptionskip}{-.7cm}
\centering
	 \subfloat[3D feature space]{
	 \includegraphics[width=0.5\linewidth]{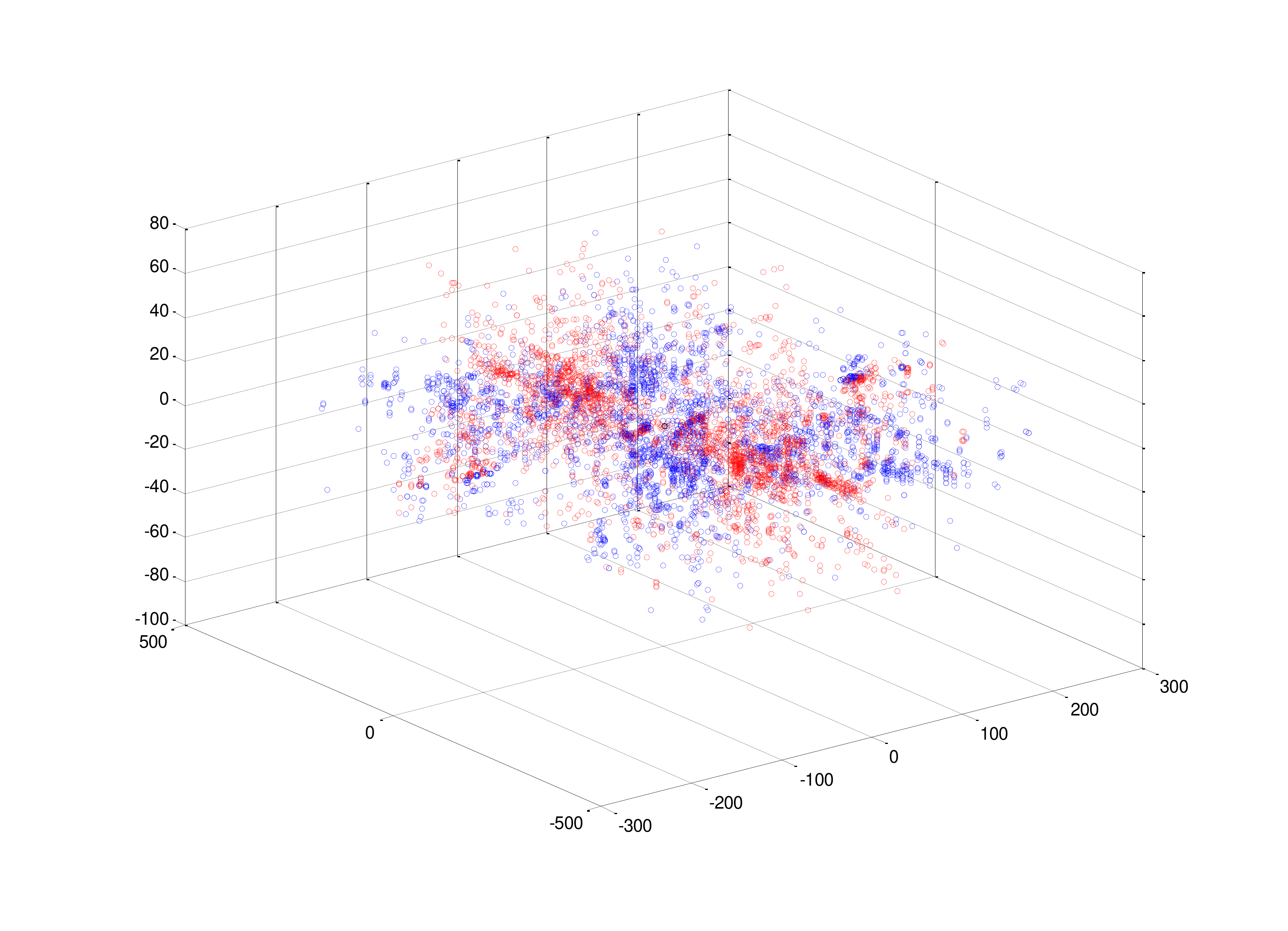}}
	 \subfloat[2D feature space]{
	 \includegraphics[width=0.46\linewidth]{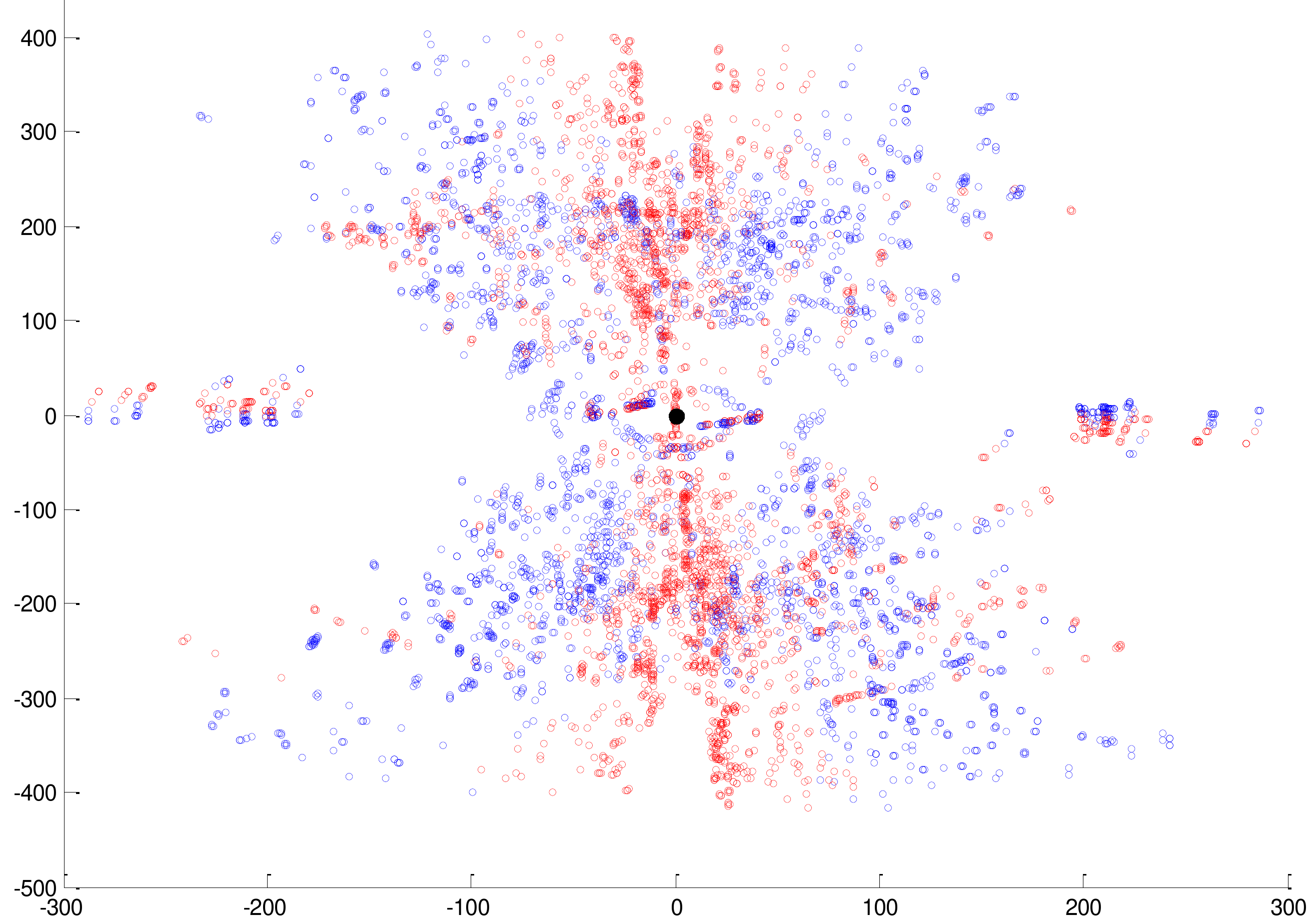}}
\caption{(a) and (b) are the  visualizations of the estimation error $\{\epsilon_n\}$ distributions in 3D and 2D feature spaces, respectively. A red dot denotes the $\epsilon$ coming from the internal learning, a blue dot represents the one from the external learning. The origin is the original HR image $X$.}
\label{fig:featuredistribution}
\end{figure}

Firstly, the distributions of $E$ produced by two learning methods are going to be explored in feature space. We selected 14 representative images $\{X_i\} (i=1,\ldots, 14)$ in SR problem and their corresponding LR images $\{Y_i\}$. The internal and external learning methods introduced in sections \ref{set:4.1} and \ref{set:4.2} are applied on each $Y_i$ by varying the configuration settings (e.g. different training datasets, image rotations and matching criteria), and result in 10 super resolved images $\{\hat{X}_i^j\} (j=1,\ldots, 10)$ where five come from the external learning method and the other are from the internal learning method, respectively. In total, there are 140 different SR images. We further compute the estimation error map $E_i^j = (\hat{X}_i^j - X_i)$ and slid an $80\times 80$ window on it to obtain a set of sub-images. From each error map, 50 resulted sub-images from either the external learning method or the internal learning method are randomly chosen. As a result, 7000 sub-images $\{\epsilon_n\} (n = 1, \ldots, 7000)$ are used for analysis. To visualize the attribute of $\{\epsilon_n\}$ in the feature space, they are projected into 3D and 2D spaces using the locality preserving projections (LPP) \cite{niyogi2004locality} and plotted in Fig. \ref{fig:featuredistribution}. The distribution illustrates that internal learning and external learning do provide different estimation errors (in other words, \textit{recovered details}). Particularly, the 2D distribution shows that the external learning recovers more diverse details whose distribution is sparser, but the internal learning concentrates on the specific detail recovery whose distribution is denser. One can see two methods perform less overlap in the 3D and 2D feature spaces. This observation suggests that they are complementary in the SR problem. Moreover, the $\epsilon_n$, which come from the same learning method but with varied configurations, locate differently in the feature space, which demonstrates that SR images are the interaction result of the learning methods and the training data.

\begin{figure*}[!t]
\setlength{\abovecaptionskip}{-0cm}
\setlength{\belowcaptionskip}{-.4cm}
\centering
	\subfloat[]{
	\includegraphics[height=.282\linewidth, width=.14\linewidth]{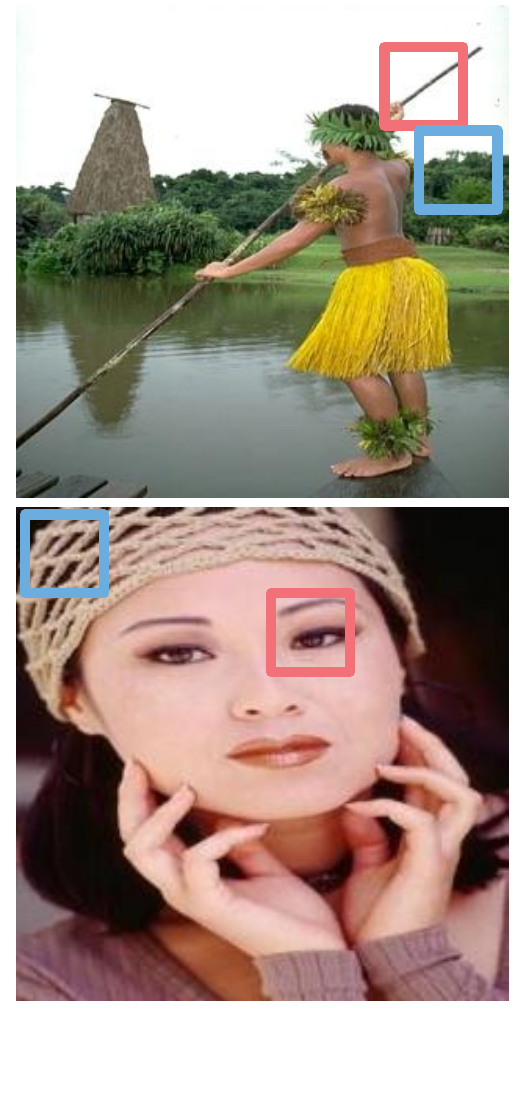}}
	\subfloat[]{
	\includegraphics[width=.37\linewidth]{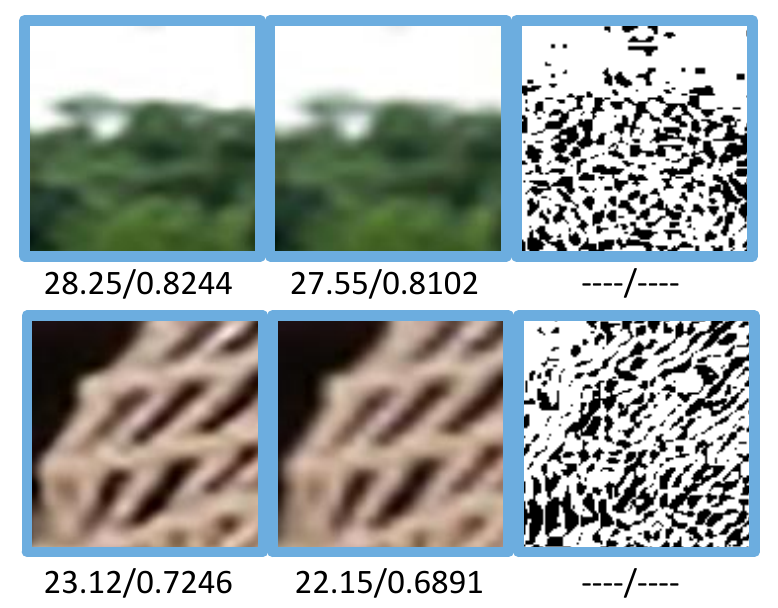}}
	\subfloat[]{
	\includegraphics[width=.37\linewidth]{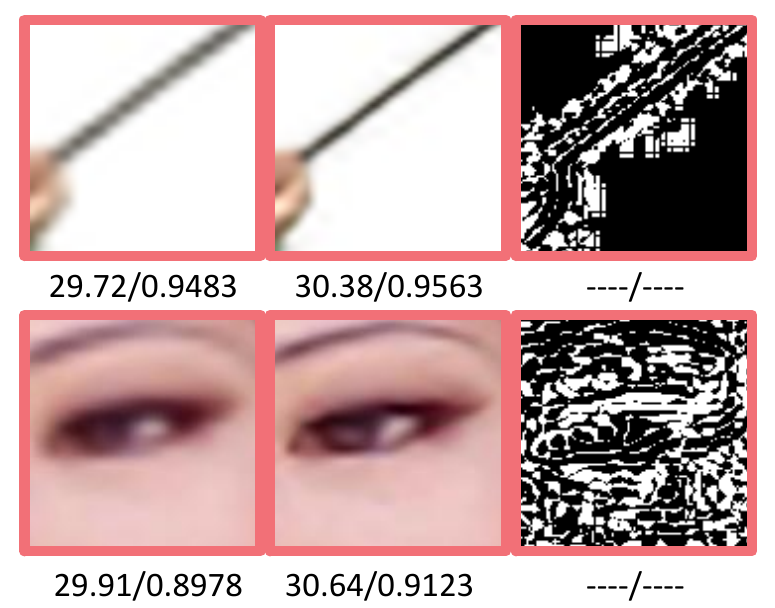}}
\caption{(a) Two image samples: \textit{Person} and \textit{Woman}. (b) The patches where the internal learning outperforms the external learning (\textit{left}. two cropped patches from the internal learning result, \textit{middle}. two cropped patches from the external learning result, \textit{right}. the preference maps, \textit{bottom}. PSNR / SSIM of this patch). (c) The patches where the external learning outperforms the internal learning (\textit{left}. results from the internal learning, \textit{middle}. results from the external learning, \textit{right}. the preference maps, \textit{bottom}. PSNR / SSIM of this patch). }
\label{fig:preferencemap}
\end{figure*}

Secondly, we explore the $E$ distribution on the image plane produced by two learning methods. To illustrate the difference, we produced a \textit{preference map} (PM) that visualizes which method would benefit a pixel in the HR image. Meanwhile, we also inspect the overlap between two error groups by counting the quantity of the same errors. Specifically, given a LR image, we calculate
\begin{equation}
\begin{split}
&|E|_{int} = abs(\hat{X}_{int} - X), \quad |E|_{ext} = abs(\hat{X}_{ext} - X),\\
&C_{int} = card(|E|_{int} > t), \:\: \:\:\: C_{ext} = card(|E|_{ext} > t), \\
&C_{overlap} = card(|E|_{int} \& |E|_{ext} > t),
\end{split}
\label{eq:error-overlap}
\end{equation}
where $\hat{X}_{int}$ is the internal learning result, $\hat{X}_{ext}$ is the external learning result. $abs(\cdot)$ denotes the element-wise absolute value in a matrix. $card(\cdot)$ is a function of counting the number of non-zero elements in a matrix. $C_{int}$ and $C_{ext}$ denote the numbers of elements whose values are greater than a threshold $t$ in the absolute error maps $|E|_{int}, |E|_{ext}$ respectively, $C_{overlap}$ denotes the number of above elements at the same locations in two error maps. Here, setting $t=7$ is to emphasize the high value components in error map that largely correspond to those high frequency signals in the SR result, because recovering the high frequency signals is the major concern of a SR algorithm.

Each pixel $\text{PM}(p,q)$, which locates at $(p,q)$ in PM, is assigned either 0 (black) or 255 (white) by comparing its corresponding $|E|_{int}(p,q)$ and $|E|_{ext}(p,q)$,
\begin{equation}
\text{PM}(p,q) =
	\begin{cases}
	255          & \text{if } \quad |E|_{int}(p,q) < |E|_{ext}(p,q)  \\
	0      & \text{if }  \quad otherwise.
	\end{cases}
\label{eq2}
\end{equation}

\begin{figure}[!t]
\setlength{\abovecaptionskip}{-0cm}
\setlength{\belowcaptionskip}{-.4cm}
\centering
\subfloat[]{
\includegraphics[width=.85\linewidth]{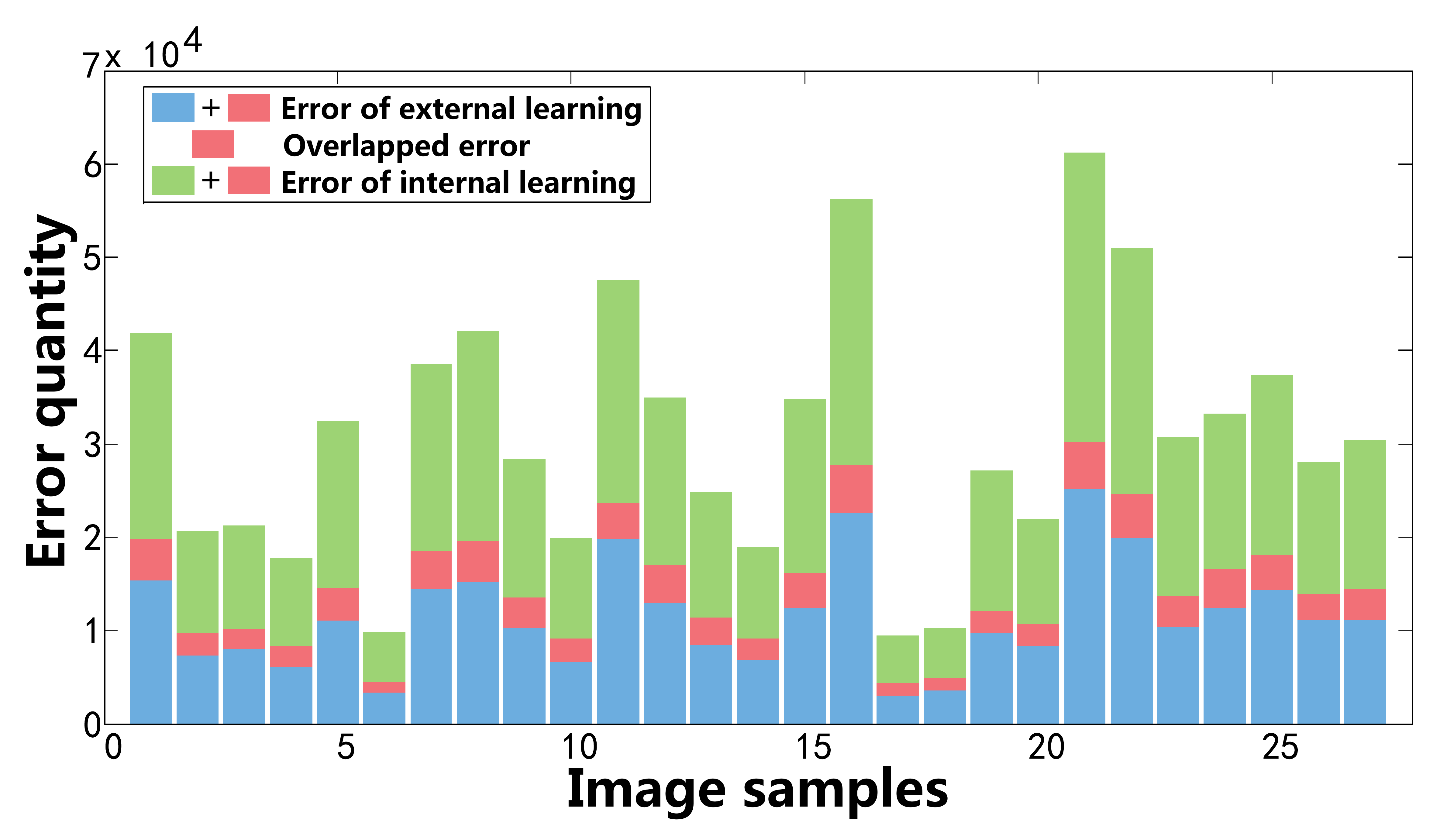}}\\
\subfloat[]{
\includegraphics[width=.9\linewidth]{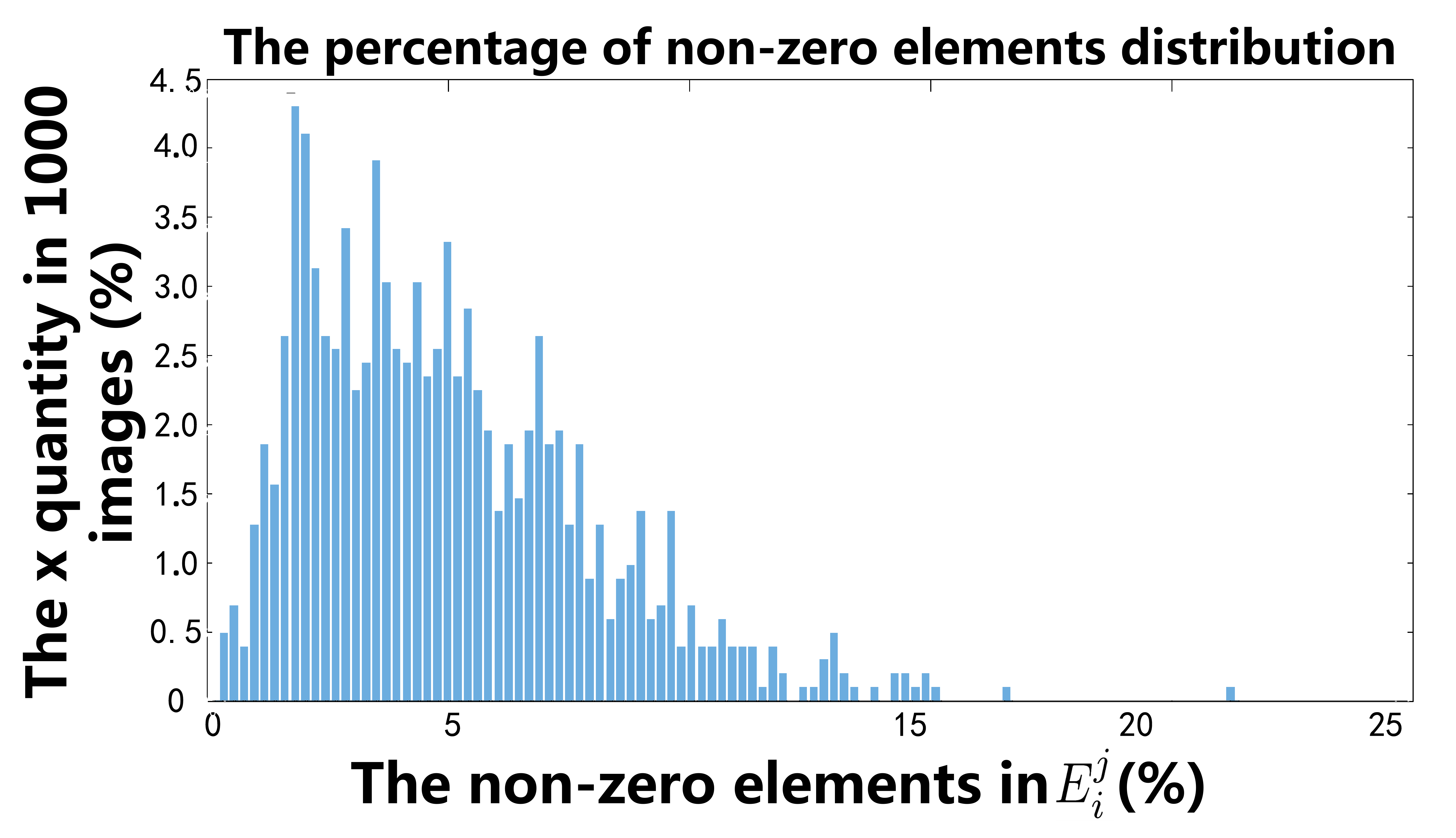}}
\caption{(a) The quantities of estimation errors from the internal learning and the external learning, and their overlap. (b) The error sparsity histogram of all 1000 error maps. The $x$ axis is the percentage of non-zero elements in one $|E|_i^j$, and $y$ axis is the percentage of corresponding $x$ quantity in 1000 samples.}
\label{fig:overlaperror}
\end{figure}

Fig. \ref{fig:preferencemap}  demonstrates two samples, \textit{Person} and \textit{Woman}. The preference maps visually show that the internal learning does perform better on the repeated patterns in the input LR image due to the self-similarity, e.g. the hair nets \& bush in Fig. \ref{fig:preferencemap}.(b), and the external learning has better recovery on those details commonly shared in the training datasets, see Fig. \ref{fig:preferencemap}.(c). To have a quantitative validation, Fig. \ref{fig:overlaperror}.(a) plots the stacked bar of three groups according to the calculation Eq. (\ref{eq:error-overlap}), where the red bar denotes the errors produced by both methods, the green bar is the errors which resulted only from internal learning, analogously, external learning is the blue bar. This statistic reveals the above fact again, $C_{overlap} \ll C_{int}$ and $C_{overlap} \ll C_{ext}$.  Both qualitative and quantitative assessments conclude that the internal and external learnings are complementary for SR problem on the image plane as well.

Till now, one can see that internal learning and external learning are complementary in both feature space and image plane, and could further improve the quality of SR image if wisely integrated.


\subsection{Sparsity Analysis}
\label{sec:3.2}
Except for the complementarity, we also analyze the intensity of $E$ that inspires us a wise integration. We randomly selected 100 LR images $\{Y_i\}$ from BSD300 database \cite{martin2001database}, then applied the internal and external learning methods on each $Y_i$ as the same in the complementarity analysis. In total, 1000 varied HR images $\{\hat{X}_i^j\}$ are obtained. Again, we computed each absolute estimation error map $|E|_i^j = abs(\hat{X}_i^j - X_i)$ and then counted the number of non-zero elements. In this study, we only consider the pixels whose error is greater than a threshold $t$. Finally, an overall statistical histogram, Fig. \ref{fig:overlaperror}.(b), plots the percentage of non-zero elements in $|E|_i^j$ among these 1000 samples. One can see that most of the samples ($>95\%$) are composed of very few elements with significant values. This reveals the major estimation errors of both learning methods are sparse.

\section{The Low-Rank Solution for Integration}
\label{sec:4}
From the above analyses of attribute, we have two observations: 1). Given an input, the internal learning and the external learning produce similar HR images with different but complementary recovered details. These high-dimensional results (HR images) share a common low-dimensional structure because they look similar. 2). The estimation errors of both methods are sparse. These observations inspire us to develop a low-rank matrix decomposition solution to integrate the pros of two methods. However, the conventional SR methods usually produce one result. Only two inputs, one from the internal learning and another from the external learning, cannot take advantage of the low-rank solution. Therefore, we tailor the configuration settings for the internal learning and the external  learning to have a bank of preliminary HR images, which are diverse and have essential and complementary priors acquired for the final SR images reconstruction. Afterward, a parameter-free low-rank solution is proposed to combine the pros of both internal and external learning methods according to their inherent property. Fig. \ref{fig:pipeline} illustrates the pipeline of the proposed framework.

\subsection{A Tailored Internal Learning}
\label{set:4.1}
The internal learning is in light of a finding that small patches in nature images often recur within and cross scale spaces \cite{freedman2011image}. Given an input LR image $Y$, enlarge it to $Y'$ with zooming factor, meanwhile, use a high-pass filter to decompose $Y$ into the low frequency band $Y^l$ and the high frequency band $Y^h$. Then, extract patches $y'_i, i=1,...,N$ from pre-enlarged image $Y'$, which act as the low-frequency component of recovered SR image patches $\hat{x}_i, i=1,...,N$. We search the most similar low-frequency patch $y^l_j, j=1,...M$ from $Y^l$ with patch $y'_i$. Thus, the corresponding high-frequency patch $y^h_j, j=1,...M$ is regarded as the high-frequency component of recovered SR image patch $\hat{x}_i, i=1,...,N$. The SR patch $\hat{x}_i$ is reconstructed by $\hat{x}_i = y'_i + y^h_j$. We keep enlarging $\hat{x}_i$ with zooming factor and repeat the above process until the desired size of SR image is reached.


To alleviate the ``Less diversity'' problem, we attempt to produce multiple different HR images. By varying the number of similar patches to be matched ($k$), the generated preliminary HR images have certain variation to ensure the diversity. However, experiments show the quantity of similar patches in an image is limited. If $k$ is set to a large value, some selected patches would have considerable differences and then result in an unfaithful result. To expand the patch searching space, we geometrically deform the ``translated'' patches by estimating an affine transformation matrix that maps the target patch to its nearest neighbors in the downsampled image. By doing this, a feasible range of $k$ value can be found to ensure both accuracy and diversity. With $k$ ready, we follow the work \cite{buades2005non} to compute the construction weight $w_{i,j}$ of each matched patch by minimizing
\begin{equation}
    \label{6}
w_{i,j} = \frac{1}{Z(i)} \exp(- \frac{||y_i - y_j||^2_2}{h^2}),
\end{equation}
where $Z(i)$ is the normalizing constant and the parameter $h$ is a filter degree parameter. Second, our patch matching is based on the $\ell_2$-norm (Euclidean distance), which is sensitive to rotation. We find that matching under varied angles results in different results. The similar observation is also reported in \cite{timofte2016seven}. Thus, we put the LR image at four angles ($0^{\circ}, 90^{\circ}, 180^{\circ}, 270^{\circ}$), and search the similar patches in each rotated image. Finally, we are able to obtain $k \times 4 $ preliminary HR results from the internal learning method.

\subsection{A Tailored External Learning}
\label{set:4.2}
External learning infers the missing high-frequency details in LR images by training a sparse representation on external datasets. It suggests that image patches can be well represented as a sparse linear combination of atoms via an appropriate over-complete dictionary. By jointly training LR-HR dictionaries, we are able to obtain the relationship of sparse representations between the low- and high-resolution image patches. To have SR results for test images, the sparse coefficients of LR image patches are firstly computed, and then the HR patch coefficients can be estimated through the obtained relationships between LR-HR patches. In the end, the HR patches' pixels are derived by multiplying the HR dictionary.

Similar to internal learning, we also expect multiple variant HR outputs to ensure the diversity of external learning. To this end, we build up multiple LR-HR dictionaries from different external training datasets. However, traditional coupled dictionary learning method is a bi-level optimization problem, which is good at learning the common characteristic from training databases but ignores the specific ones. Domke et al. \cite{Domke12} reported that bi-level optimization suffers from high computation cost in training and is difficult to reach the specific characteristic because of massive iterations of minimizing the internal function. His solution truncates the internal optimization to a fixed number of iterations. By back optimizing the internal parameters, a better result can be achieved which has more specific characteristics from each training dataset. Inspired by \cite{Domke12}, we truncate the sparse encoding procedure ISTA \cite{beck2009fast} to $i$ iterations in LR dictionary training. We also find the truncated algorithm meets our criterion even when $i = 1$. The encoding function is then formulated as:
\begin{equation}
    \label{9}
z  = {S_\theta }\left( {{W_e}y + W_d{S_\theta }\left( {{W_e}y} \right)} \right),
\end{equation}
\noindent where $z$ is the sparse representation, $y$ is the LR image patches,  $S_\theta$ is a soft thresholding function, $\theta$ is a vector of threshold, $W_e$, $W_d$ are equivalent to $W^T$, $W^T W$, respectively,  $W$ is a prespecifie LR dictionary in the ISTA algorithm, here $D_l = \left\{ {{\theta};{W_e};{W_d}} \right\}$ is the learned LR dictionary. For more details on the parameter $D_l$, please refer to \cite{gregor2010learning}.

Given $m$ varied training datasets, each dictionary-pair is computed by
\begin{equation}
\begin{aligned}
&D_{h},D_{l} = \argmin \Vert {{x} - {D_{h}}z} \Vert_F^2, \\
&\textrm{subject to:}  \\
&{\begin{cases}
    z  = {S_\theta }\left( {{W_e}y + W_d{S_\theta }\left( {{W_e}y} \right)} \right),\\
    D_l = \left\{ {{\theta};{W_e};{W_d}} \right\},\\
\end{cases}}
\end{aligned}\label{eq3}
\end{equation}
\noindent where $x$ denotes HR image patches, $\Vert {\cdot} \Vert_F^2$ denotes the Frobenius norm.

Because image patch in dictionary learning is represented by a vector, it is sensitive to patch rotation and results in different SR results with changes in angles.  Similar to the tailored internal learning, we put the LR input at four angles ($0^{\circ}, 90^{\circ}, 180^{\circ}, 270^{\circ}$) and apply our external learning method on each. Then, we can have $m \times 4$ preliminary HR results.

\subsection{The Low-Rank Solution}

\subsubsection{The Low-Rank Matrix Decomposition Model}
\label{set:4.3.1}
From above tailored internal and external learning methods, multiple preliminary HR images are reconstructed from the self-similarities and the learned meta-detail. They have the essential but different recovered details that can be considered as new priors. It is natural to use them together to further improve the result. However, each image also has its own estimation errors and noise beside the prior. Thus, a straightforward integration (e.g. taking a weighted average) does not work. Due to the significant correlation among these HR images, we vectorize and then stack them up as a matrix $X$. The correlated components in $X$ can be reduced into a low-dimensional subspace, while the uncorrelated components (estimation errors and noise) are left in the original space. In other words, $X$ can be approximated by a low-rank matrix. We, therefore, treat the integration as a dimension reduction problem, and propose to decompose $X$ into a low-rank component and a sparse component. The reason is that low-rank decomposition was designed for extracting correlation, and has proven its robustness and effectiveness on separating the low-rank component from the sparse component in the data. In this problem, the low-rank component is the desired SR image, and the sparse component contains the estimation errors and noise.


To obtain the low-rank component, we let $X=L+N$, where $L$ denotes the low rank matrix, $N$ is the perturbation matrix. The smaller the $N$ is, the better performance we achieve. Thus, the problem now becomes how to find the best $r$ rank estimation of the matrix $L$, and is formulated as:
\begin{equation}
\min\limits_L\|X-L\|\quad s.t. \quad rank(L)\leq r.
\label{eq8}
\end{equation}

Instead of perturbation matrix $N$, low-rank decomposition uses sparse representation $S$. So, $X$ is further written as $X=L+S$. $L$ and $S$ are computed by
\begin{equation}
\min\limits_{L,S}\|L\|_*+\lambda\|S\|_1\quad s.t.\quad L+S=X,
\label{eq9}
\end{equation}
where $\|L\|_*$ is the nuclear norm of the matrix $L$ (the sum of singular values of $L$), $\|S\|_1$ is the $\ell_1$-norm of $S$. Because of huge computation of SVD decomposition of large matrix and the problematic under-determinate objective (\ref{eq9}), we employ an alternating projection algorithm that is like the Go-Dec method \cite{zhou2011godec}. Specifically, we alternately compute one by fixing another one as follows:
\begin{equation}
\begin{split}
&L_t=\mathop {\arg \min }\limits_{rank(L)\leq r}\;\|X-L-S_{t-1}\|_F^2,\\
&S_t=\mathop {\arg \min }\limits_{card(S)\leq k}\;\|X-L_t-S\|_F^2,
\end{split}
\label{eq10}
\end{equation}
\noindent where $card(\cdot)$ denotes the number of non-zero elements in the matrix.

In this problem, the rank $r$ is expected to be 1. However, existing low-rank algorithms cannot guarantee a global solution with $r=1$. Alternatively, the average of the low-rank components (images) can be the ultimate SR result. As the LR images often involve noise and they are random in real world scenarios, our algorithm also calculates the noise matrix $G$ when computing $S$. This shows another capability, anti-noise, which means our solution is able to super-resolve the noisy images. Experiments in section \ref{set:5.4} demonstrate this additional advantage.




\subsubsection{Input Quantity vs. Performance}
\label{set:4.3.2}
In practice, we encounter a question that how many inputs are feasible in our low-rank solution to have the desired SR result. Fortunately, some progress has been made in theories of low-rank modeling. Cand\`{e}s at al. \cite{Candes&ma, candes&plan, candes&tao} and Chandrasekaran et al. \cite{Chandrasekaran2011} have theoretically proven that the solution of (\ref{eq9}) can recover the low-rank matrix $L$ and the sparse matrix $S$ with a high probability, if the following conditions are met: 1. The underlying $L$ satisfies the incoherence condition\footnote{The incoherence condition mathematically characterizes the difficulty of recovering the underlying $L$ from a few sampled entries. Informally, it says that the singular vectors of $L$ should sufficiently ``spread out'' and be uncorrelated with the standard basis.}; 2. The non-zero entries of the underlying $S$ are sufficiently sparse with a random spatial distribution; 3. The observed entries
are uniformly distributed in matrix $X$, and the lower bound on number of them is on the order of $\mathcal{O}(nr\log n)$, where $n$ is the image size (product of width and height of an image). Firstly, our problem satisfies the condition 1. Secondly, section \ref{sec:3.2} proves the estimation error is sparse, meanwhile, the noise is random. Both meet the condition 2. Finally, the condition 3 indicates that a detail from either the internal learning or external learning is able to be recovered if it appears in $X$ more than $\mathcal{O}(r\log n)$ times. In other words, we need at least $\mathcal{O}(r\log n)$ images to recover $L$ and $S$. Commonly, $\log n$ approximates to 5 or 6 in the super-resolution problem. This means our low-rank solution is solvable when having a small number of the external and internal learning results as the input.

Unfortunately, aforementioned researches do not provide an upper bound or prove if more images will always lead to better results. To answer this question, it is better to explore the influence of $L, S$ and $G$ to the convergence of our low-rank solution. We hereinafter employ the theory founded by Lewis \& Malick \cite{Lewis2008}, then give an analysis below by considering $L$ first and with the core mathematical background.

The algorithm solving (\ref{eq10}) can be imagined as projecting $L$ onto a manifold $\mathcal{M}$ and then onto another one $\mathcal{N}$ alternately, where $\mathcal{M}$ and $\mathcal{N}$ are two $C^k$ manifolds around a point $\overline{L} \in \mathcal{M}\cap\mathcal{N}$.
\begin{equation}
\left\{
\begin{split}
&\mathcal{M}=\{M\in\mathbb{R}^{m\times n}: rank(M)=r  \}\\
&\mathcal{N}=\{X- P_\Omega(X-M):  M\in\mathbb{R}^{m\times n}\}
\end{split}
\right. ,
\label{eq11}
\end{equation}
where $M$ is a matrix, $P_\Omega(\cdot)$ is a projection of a matrix to a set $\Omega$. We can see any point $\overline{L} \in \mathcal{M}\cap\mathcal{N}$ is a local result of Eq.(\ref{eq10}) because
\begin{equation}
\begin{split}
S = & P_\Omega(X-L),  \\
L = & P_{\mathcal{M}\cap\mathcal{N}}(L) = X - S =X- P_\Omega(X-L), \\
& s.t. \quad rank(L)=r.
\end{split}
\label{eq12}
\end{equation}

Thus, updating $L_{t+1}$ in Eq.(\ref{eq10}) is achieved by
\begin{equation}
\begin{split}
L_{t+1} & =  P_\mathcal{M}(X-S_t) \\
 & = P_\mathcal{M}(X - P_\Omega(X-L_t)) = P_\mathcal{M}P_\mathcal{N}(L_t).
\end{split}
\label{eq13}
\end{equation}

The theory \cite{Lewis2008} states that a smaller angle between these two manifolds produces a faster convergence and better result of the algorithm. Therefore, our task becomes to discuss how $L, S$ and $G$ influence the angle between two manifolds $\mathcal{M}$ and $\mathcal{N}$ in $\mathbb{R}^{m\times n}$ which is defined as
\begin{equation}
\begin{split}
c(\mathcal{M}, \mathcal{N}) = max \left\{  \langle x,y\rangle: x\in \mathbb{S}\cap \mathcal{M}\cap (\mathcal{M} \cap\mathcal{N})^\perp, \right. & \\
\left. y\in \mathbb{S}\cap \mathcal{M}\cap (\mathcal{M} \cap\mathcal{N})^\perp\right\}, &
\end{split}
\label{eq14}
\end{equation}
where $\langle \cdot\rangle$ is the inner product, $\mathbb{S}$ is the unit sphere in $\mathbb{R}^{m\times n}$. The angle at point $\overline{L}$ is defined as the angle between the corresponding tangent spaces $T_\mathcal{M}(\overline{L})$ and $T_\mathcal{N}(\overline{L})$.
$$c(\mathcal{M}, \mathcal{N}, \overline{L})) = c(T_\mathcal{M}(\overline{L}), T_\mathcal{N}(\overline{L})).$$

The normal spaces of manifolds $\mathcal{M}$ and $\mathcal{N}$ at point $\overline{L}$ are given by
\begin{equation}
\begin{split}
N_\mathcal{M}(\overline{L}) &=\{y\in\mathbb{R}^{m\times n}:u_i^Tyv_j = 0,  \overline{L} = UDV^T \},\\
N_\mathcal{N}(\overline{L}) &=\{X- P_\Omega(X-\overline{L})\},
\end{split}
\label{eq15}
\end{equation}
where $UDV^T$ is the eigenvalue decomposition. Assume $X = \overline{L}+\overline{S}+\overline{G}$, we have
\begin{equation}
\begin{split}
\overline{L} & =X - (\overline{S} + \overline{G}), \\
\hat{L} & =X - P_\Omega(\overline{S} + \overline{G}), \\
& = \overline{L} + [(\overline{S} + \overline{G}) - P_\Omega(\overline{S} + \overline{G})] = \overline{L} + \Delta_L.
\end{split}
\label{eq16}
\end{equation}
where $\overline{G}$ is the noise corresponding to $\overline{L}$, $\Delta_L$ is the element-wise hard thresholding error of $\overline{S} + \overline{G}$. Thus, the normal space of manifold of $\mathcal{N}$ is
\begin{equation}
N_\mathcal{N}(\overline{L}) = \{\overline{L} + \Delta_L\}.
\label{eq17}
\end{equation}

Due to the tangent space is the complement space of the normal space, we have
\begin{equation}
N_\mathcal{N}(\overline{L}) \subseteq T_\mathcal{M}(\overline{L}), \qquad N_\mathcal{M}(\overline{L}) \subseteq T_\mathcal{N}(\overline{L}).
\label{eq18}
\end{equation}

Therefore, the angle at point $\overline{L}$ can be simplified as
\begin{equation}
\begin{split}
c(\mathcal{M}, \mathcal{N}, \overline{L}) = max \left\{  \langle x,y\rangle: x\in \mathbb{S}\cap N_\mathcal{N}(\overline{L}), \right. & \\
\left. y\in \mathbb{S}\cap N_\mathcal{M}(\overline{L})\right\} . &
\end{split}
\label{eq19}
\end{equation}

According to equations (\ref{eq15}), (\ref{eq17}) and (\ref{eq19}), the angle is
\begin{equation}
\begin{split}
\langle x,y\rangle & = tr(VDU^T y + \Delta_L^Ty) \\
 & = tr(DU^TyV) + tr(\Delta_L^Ty) =  tr(\Delta_L^Ty).
\end{split}
\label{eq20}
\end{equation}

Then, we can see
\begin{equation}
c(\mathcal{M}, \mathcal{N}, \overline{L}) = max \left\{\langle x,y\rangle \right\}\leq max\left\{\langle D(\Delta_L), D(y)\rangle \right\},
\end{equation}
where the diagonal elements of $D(\Delta_L)$ and $D(y)$ are eigenvalues of $\Delta$ and $y$, respectively. Moreover, $\overline{L} + \Delta_L, y\in \mathbb{S}$ implicates that $\Vert \overline{L} + \Delta_L\Vert_F^2 = \Vert y \Vert_F^2 = 1$, the above inequality can be deduced as
\begin{equation}
\begin{split}
c(\mathcal{M}, \mathcal{N}, \overline{L}) & \leq max\left\{\langle D(\Delta_L), D(y)\rangle \right\} \\
& \leq \Vert D(\Delta_L) \Vert_F\Vert D(y)\Vert_F \leq \Vert D(\Delta_L) \Vert_F.
\end{split}
\end{equation}

Till now, we can conclude that the angle at a local solution $\overline{L}$ is influenced by the $\Vert D(\Delta_L) \Vert_F$. In other words, the convergence of $L$ will slow down and the resulting algorithm will degrade while $\Vert\Delta_L\Vert_F$ is augmented.

For $S$, the analogous analysis can be done as above, in which $\Delta_S = (L+ G) - P_\mathcal{M}(L+G)$ is the singular value hard thresholding error of $L + G$.

The above analysis theoretically indicates that a better performance of our low-rank solution can be achieved by abating $\Vert\Delta \Vert_F$. As aforementioned in sections \ref{set:4.1} and \ref{set:4.2}, we increase the diversity of internal and external learnings to produce multiple preliminary HR results as the input of the proposed low-rank solution. It aims at increasing the appropriate details from self similarities and external samples, and integrating them as much as possible. However on the other hand, too many preliminary HR results augment $\Vert\Delta \Vert_F$ of both $L$ and $S$ as well. One can see the relation between the low-rank input quantity and its performance is not monotonically increasing. When the quantity is small, the learned details from new entries contribute to the SR image more than the associated error $\Vert\Delta \Vert_F$ does. Then the performance improves with adding more preliminary HR images. Nevertheless, massive input entries will not bring significant new details into the low-rank solution further, but introduce more errors which degrade SR image quality and slow down the convergence. In this case, the quality of the resulted SR image gets worse when the preliminary HR images are sustainedly added. Ideally, we expect to find the turning point which reaches at the peak of the performance against the input quantity. Our experiments show that only around 26 $\sim$ 42 preliminary HR images achieve the best performance, please refer to Fig. \ref{quantity_performance}. This conclusion allows a much simpler tailoring of the internal learning and external learning to fit the low-rank solution.

\begin{figure*}[!t]
\setlength{\abovecaptionskip}{-0cm}
\setlength{\belowcaptionskip}{-.4cm}
\centering
    \includegraphics[width=1\linewidth]{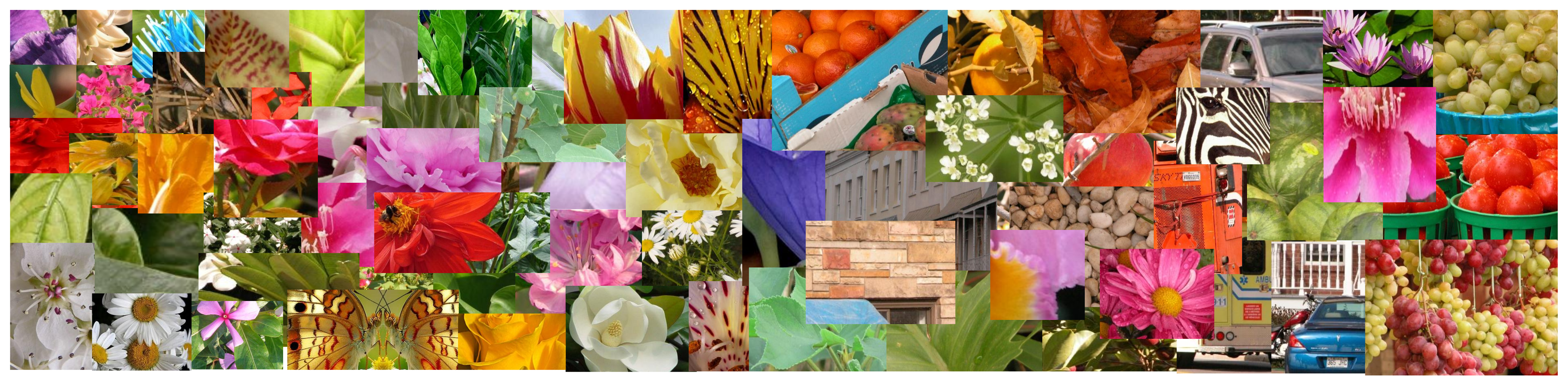}
\caption{The samples of training images.}
    \label{example_images}
\end{figure*}

\section{Experiments and Discussion}
\label{sec:5}
In order to demonstrate the effectiveness of our proposed method, the qualitative and quantitative assessments are conducted on five aspects: 1). the experiment of low-rank input quantity vs. its performance, 2). comparison with recent state-of-the-arts on noiseless images, 3). comparison on the synthetic noisy images with varied noises, 4). comparison on the real noisy images, and 5). comparison with denoising + super-resolution on noisy images.

The comparison baselines consist of six state-of-the-art methods: the adjusted anchored neighbor neighbor regression (A+) \cite{timofte2014a+}, the transformed self-example (TSE) \cite{huang2015single} and the deep learning based SR methods: the Super-resolution Convolutional Neural Network (SRCNN) \cite{dong2014learning}, the Sparse Coding based Network (SCN) \cite{liu2016robust}, Fast Super-Resolution Convolutional Neural Network (FSRCNN) \cite{dong2016accelerating} and Very Deep Convolutional Networks for Image Super-Resolution (VDSR) \cite{Kim_2016_VDSR}, where TSE is an internal learning method, others are external learning methods, particularly, SCN can super resolve noisy images. In addition, we also assess the low-rank solution on internal learning / external learning individually, named Internal$_\text{LRS}$ / External$_\text{LRS}$. The quantitative assessments are carried out on two evaluation metrics, namely peak-signal-to-noise ratio (PSNR) and structural similarity (SSIM). PSNR is the ratio between the reference signal and the distortion signal in an image. SSIM measures the structure changes between the reference and distorted image, which stresses the topology information.

\subsection{Experimental Configuration}
\label{sec:5.1}

To train the external learning method, 4.1 million image patch-pairs are randomly selected from database \cite{yang2010image} where the LR images are upsampled to the same size as the corresponding HR ones and then both of them are partitioned into patches of size $9 \times 9$. Next, the LR-HR image patch-pairs are sorted in a descending order by the patch variance, and sequentially divided into $m$ groups with an overlap of 100,000 samples between each other. Each group contributes to one dictionary in external learning. To effectively extract the high frequency signals in LR patches, the first and second order derivatives of the image patches along $x$ and $y$ axes are used as the feature (preprocessing) with dimension of 324 ($4 \times 9 \times 9$). Since it is time consuming, we then reduce the dimensionality to 30 using PCA, which preserves 99.9\% of average energy.

For the internal learning, we select varied similar patches (the number from $1$ to $k$) as self-similarity from four rotated images to reconstruct multiple HR images. Since our human visual system is more sensitive to the luminance channel than the chromatic channels, the tested images are transformed from RGB to YCbCr color space. So, the tailored internal and external learning methods perform only on the Y channel. The evaluations are carried on the noiseless images from three standard test datasets $\{$Set5 \cite{bevilacqua2012low}, Set14 \cite{zeyde2012single}, BSD100 (a set of 100 images from BSD300) \cite{martin2001database}$\}$, the synthetic noisy images by adding varied Gaussian noise to noiseless images, and real noisy images from dataset \cite{neatvideo}.

%

\begin{figure}[!t]
\setlength{\abovecaptionskip}{-0cm}
\setlength{\belowcaptionskip}{-.4cm}
\centering
    \includegraphics[width=.9\linewidth]{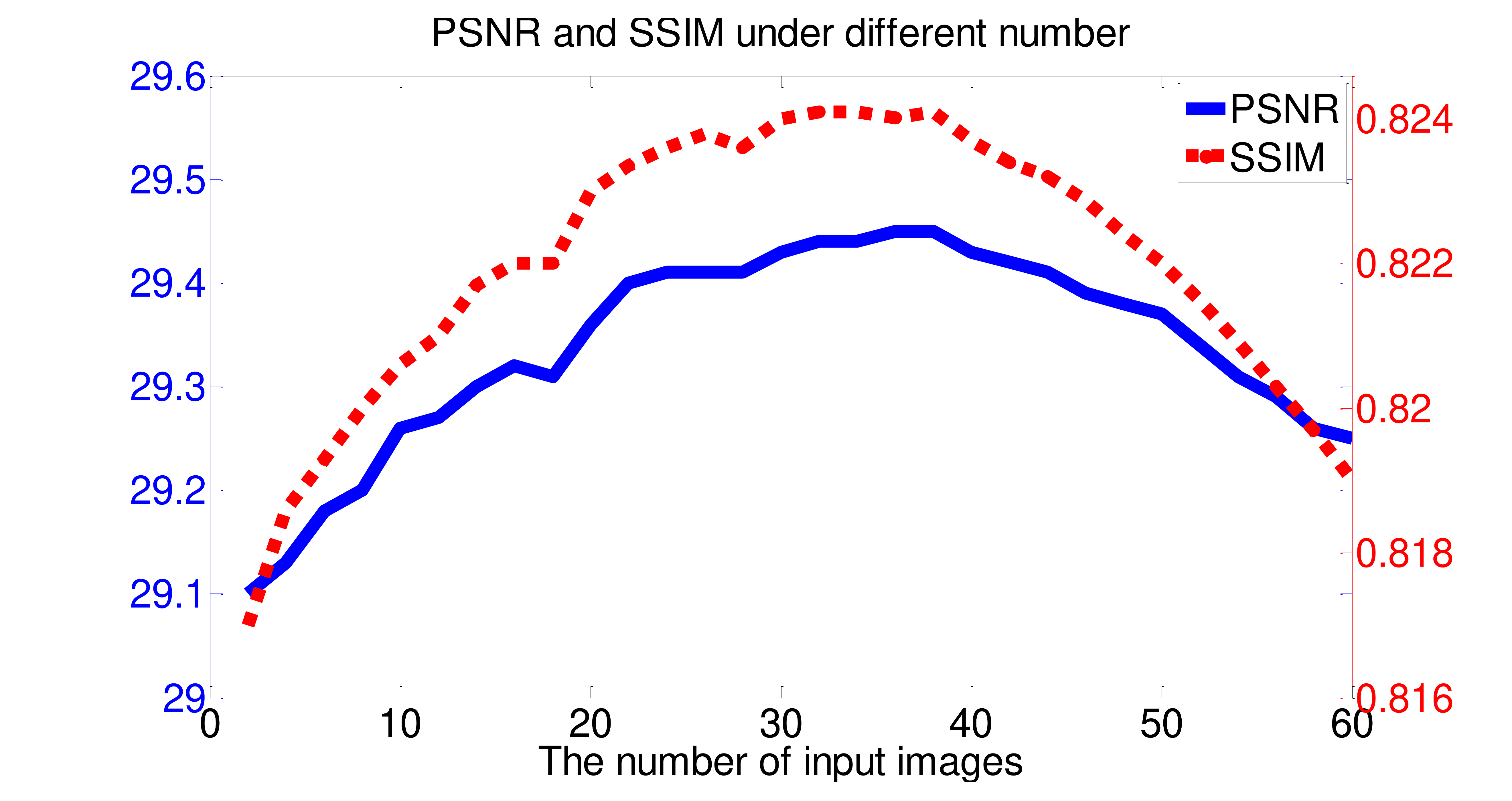}
\caption{PSNR and SSIM of proposed low-rank solution against varied numbers of input images.}
    \label{quantity_performance}
\end{figure}

\begin{figure*}[!t]
\centering
    \includegraphics[width=.9\linewidth]{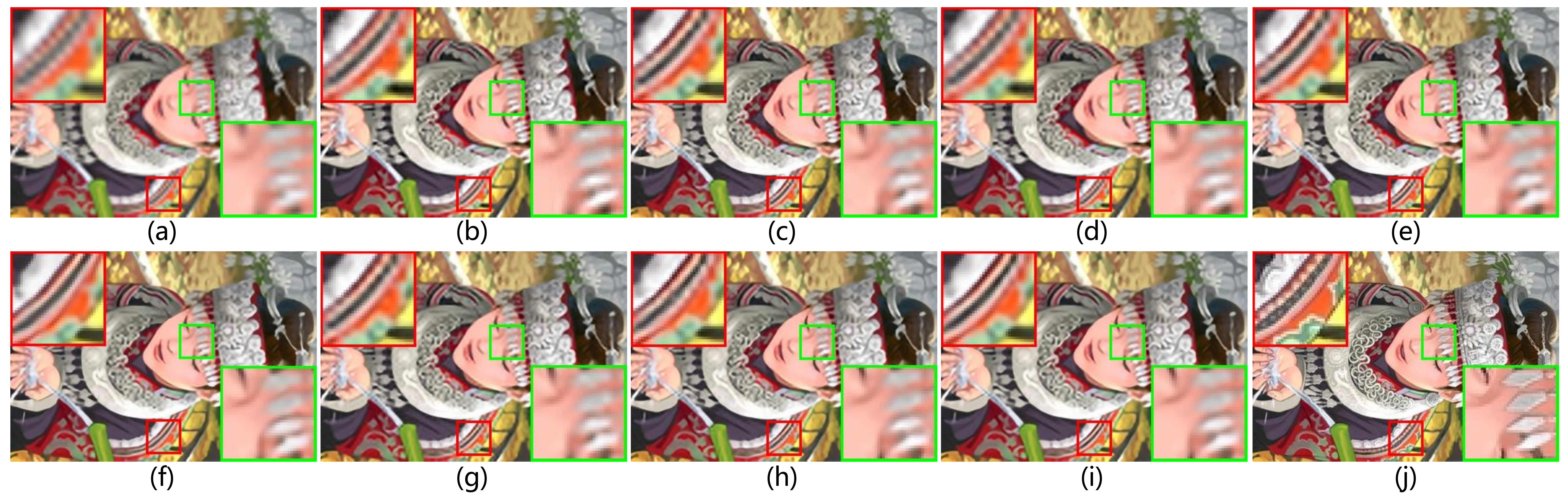}
\caption{Results of noiseless image, \emph{comic} (magnified by a factor of 3). Local magnifications are shown in the left top and right bottom of each example. (a) A+ \cite{timofte2014a+}. (b) TSE \cite{huang2015single}. (c) SRCNN \cite{dong2014learning}. (d) SCN \cite{liu2016robust}. (e) FSRCNN \cite{dong2016accelerating}. (f) VDSR \cite{Kim_2016_VDSR}. (g) External$_\text{LRS}$. (h) Internal$_\text{LRS}$. (i) Proposed method. (j) Ground truth.}
    \label{figure1}
\end{figure*}

\setlength{\belowcaptionskip}{.5cm}

\begin{table*}[t] \vspace{-4pt}
\centering
\caption{PSNR (SSIM) comparison on three standard data sets among six competing methods, External, Internal, External$_\text{LRS}$, Internal$_\text{LRS}$, and Proposed method.}
\label{tab:table1}
\vspace{1pt}
\small
\begin{tabular}{r||c|c|c||c|c|c||c|c|c}
\hline
Data Set      & \multicolumn{3}{c||}{Set5}	& \multicolumn{3}{c||}{Set14}	& \multicolumn{3}{c}{BSD100} \\
\hline
Upscaling &	$\times2$ &	$\times3$ & $\times4$ &
        	$\times2$ &	$\times3$ & $\times4$ &
            $\times2$ &	$\times3$ & $\times4$ \\
\hline
\hline
\multirow{2}{*}{A+ \cite{timofte2014a+}} &
36.55 & 32.59 & 30.29 & 32.28 & 29.13 & 27.33 & 31.21 & 28.29 & 26.82 \\
& (0.9544) & (0.9088) & (0.8603) & (0.9056) & (0.8188) & (0.7491) & (0.8863) & (0.7835) & (0.7087) \\
\hline
\multirow{2}{*}{TSE \cite{huang2015single}} &
36.49 & 32.58 & 30.31 & 32.22 & 29.16 & 27.40 & 31.18 & 28.29 & 26.84 \\
& (0.9537) & (0.9093) & (0.8619) & (0.9034) & (0.8196) & (0.7518) & (0.8855) & (0.7840) & (0.7106) \\
\hline
\multirow{2}{*}{SRCNN \cite{dong2014learning}} &
36.66 & 32.75 & 30.49 & 32.45 & 29.30 & 27.50 & 31.36 & 28.41 & 26.90 \\
& (0.9542) & (0.9090) & (0.8628) & (0.9067) & (0.8215) & (0.7513) & (0.8879) & (0.7863) & (0.7101) \\
\hline
\multirow{2}{*}{SCN \cite{liu2016robust}} &
37.21 & 33.34 & 31.14 & 32.80 & 29.57 & 27.81 & 31.60 & 28.60 & 27.14 \\
& (0.9571) & (0.9173) & (0.8789) & (0.9101) & (0.8263) & (0.7619) & (0.8915) & (0.7905) & (0.7191) \\
\hline
\multirow{2}{*}{FSRCNN \cite{dong2016accelerating}} &
37.00 & 33.16 & 30.71 & 32.63 & 29.43 & 27.59 & 31.56 & 28.52 & 27.01 \\
& (0.9558) & (0.9140) & (0.8657) & (0.9088) & (0.8242) & (0.7535) & (0.8894) & (0.7853) & (0.7131) \\
\hline
\multirow{2}{*}{VDSR \cite{Kim_2016_VDSR}} &
\textbf{37.53} & \textbf{33.66} & \textbf{31.35} & \textbf{33.03} & \textbf{29.77} & \textbf{28.01} & \textbf{31.90} & \textbf{28.82} & \textbf{27.29} \\
& (\textbf{0.9587}) & (\textbf{0.9213}) & (\textbf{0.8838}) & (\textbf{0.9124}) & (\textbf{0.8314}) & (\textbf{0.7674}) & (\textbf{0.8960}) & (\textbf{0.7976}) & (\textbf{0.7251}) \\
\hline
\hline
\multirow{2}{*}{External} &
36.46 & 32.63 & 30.31 & 32.25 & 29.16 & 27.26 & 31.17 & 28.29 & 26.90 \\
& (0.9540) & (0.9052) & (0.8614) & (0.9052) & (0.8202) & (0.7487) & (0.8864) & (0.7836) & (0.7067) \\
\hline
\multirow{2}{*}{Internal} &
36.52 & 32.55 & 30.14 & 32.19 & 29.14 & 27.41 & 31.06 & 28.26 & 26.84 \\
& (0.9594) & (0.9082) & (0.8603) & (0.9057) &
(0.8202) & (0.7506) & (0.8815) & (0.7842) & (0.7097)\\
\hline
\multirow{2}{*}{External$_\text{LRS}$} &
36.58 & 32.71 & 30.45 & 32.41 & 29.29 & 27.41 & 31.33 & 28.38 & 27.03 \\
& (0.9561) & (0.9078) & (0.8626) & (0.9061) & (0.8212) & (0.7512) & (0.8878) & (0.7858) & (0.7083) \\
\hline
\multirow{2}{*}{Internal$_\text{LRS}$} &
36.58 & 32.51 & 30.36 & 32.36 & 29.23 & \textbf{27.53} & 31.24 & 28.47 & 27.04 \\
& (0.9622) & (0.9077) & (0.8625) & (0.9064) &
(0.8214) & (\textbf{0.7524}) & (0.8832) & (0.7877) & (0.7111)\\
\hline
Our & \textbf{36.71} & \textbf{32.81} & \textbf{30.52} & \textbf{32.53} & \textbf{29.40} & 27.42 & \textbf{31.45} & \textbf{28.58} & \textbf{27.11} \\
Improvement & (\textbf{0.9633}) & (\textbf{0.9098}) & (\textbf{0.8645}) & (\textbf{0.9098}) & (\textbf{0.8236}) & (0.7503) & (\textbf{0.8881}) & (\textbf{0.7889}) & (\textbf{0.7121}) \\
\hline
\end{tabular}
\end{table*}

\begin{figure*}[!t]
\setlength{\abovecaptionskip}{-0cm}
\setlength{\belowcaptionskip}{-.2cm}
\centering
    \includegraphics[width=.9\linewidth]{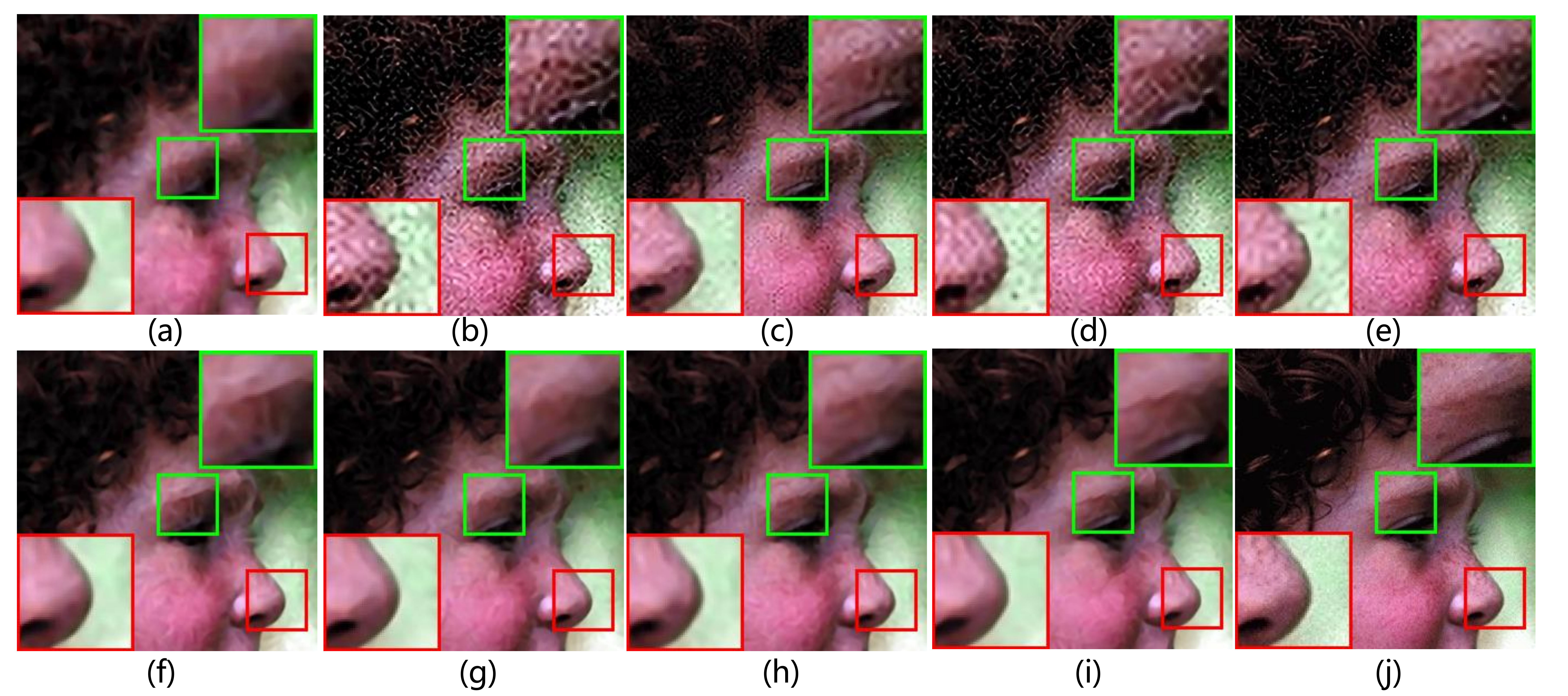}
\caption{Results of synthetic noisy image, \emph{face}, with standard deviation of 20 (magnified by a factor of 3). Local magnifications are shown in the left bottom and right top of each example. (a) A+ \cite{timofte2014a+}. (b) SRCNN \cite{dong2014learning}. (c) SCN \cite{liu2016robust}. (d) FSRCNN \cite{dong2016accelerating}. (e) VDSR \cite{Kim_2016_VDSR}. (f) TSE \cite{huang2015single}. (g) External$_\text{LRS}$. (h) Internal$_\text{LRS}$. (i) Proposed method. (j) Ground truth.}\label{noise_20}
\end{figure*}

\begin{figure*}[!t]
\setlength{\abovecaptionskip}{-0cm}
\setlength{\belowcaptionskip}{-.2cm}
\begin{center}
    \includegraphics[width=0.9\linewidth]{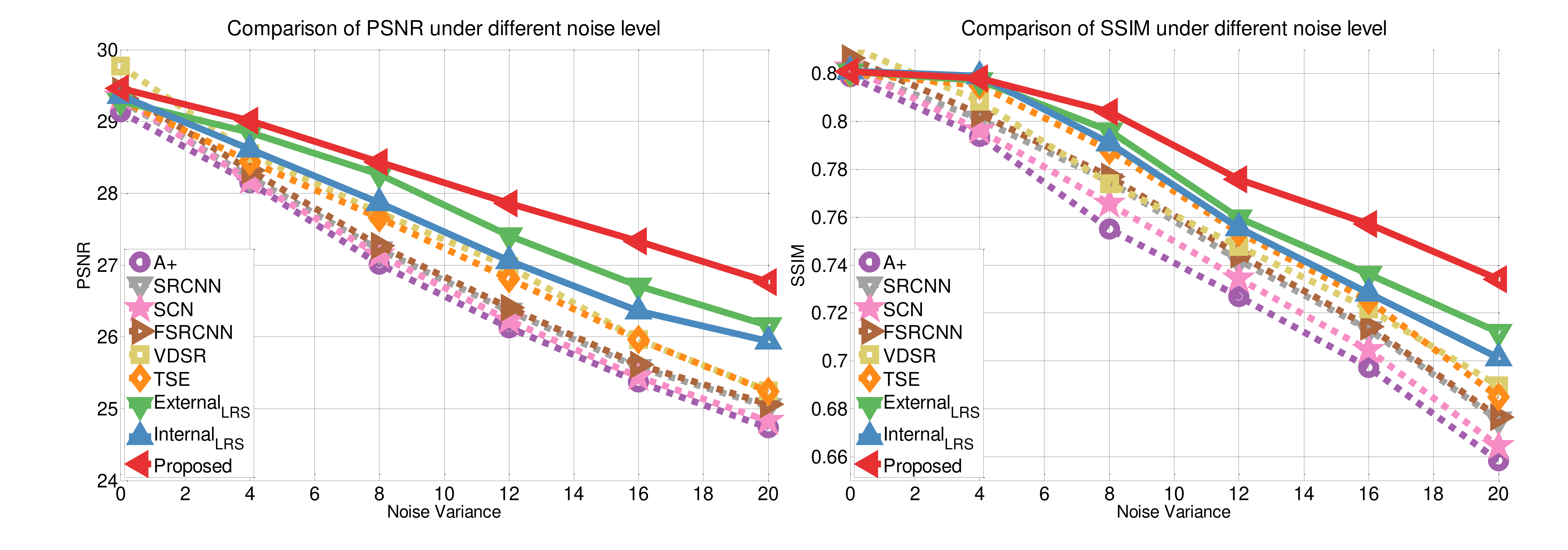}
\caption{PSNR and SSIM of six competing methods, External$_\text{LRS}$, Internal$_\text{LRS}$, and our proposed method under varied noise levels (from 4 up to 20).}
  \label{fig:noise_psnr_ssim}
\end{center}
\end{figure*}

\begin{figure*}[!t]
\setlength{\abovecaptionskip}{-0cm}
\setlength{\belowcaptionskip}{-.2cm}
\centering
    \includegraphics[width=.9\linewidth]{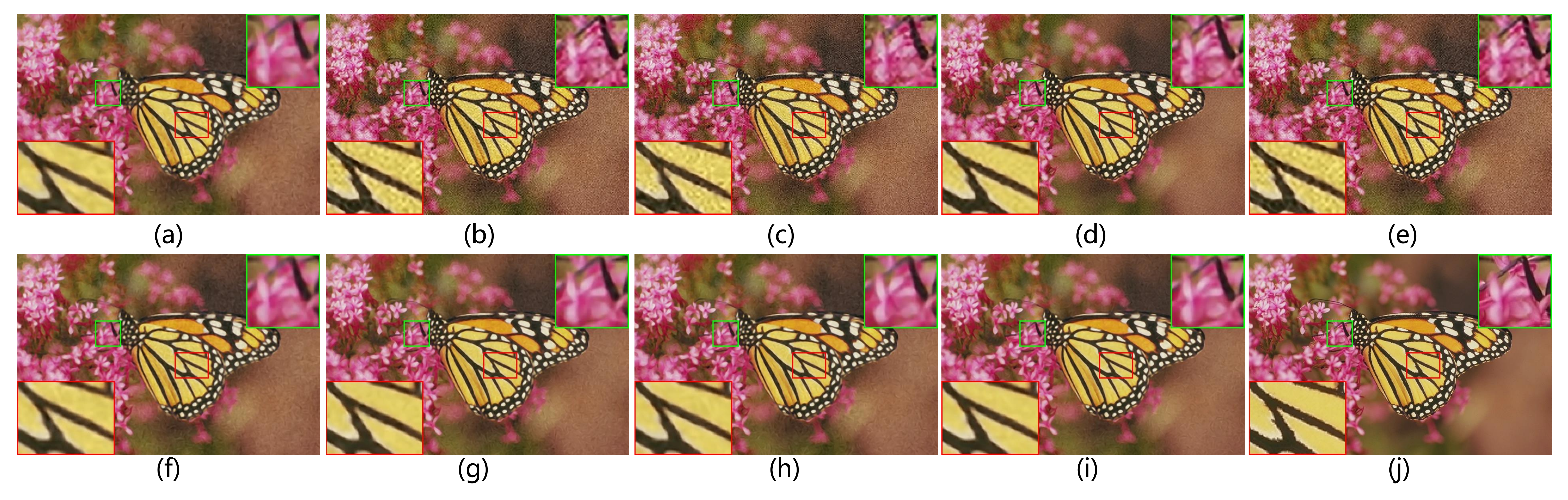}
\caption{Results of synthetic noisy image, \emph{monarch}, with standard deviation of 12 (magnified by a factor of 3). Please refer to Fig. \ref{noise_20} for the description of sub-figures.}
\label{noise_12}
\end{figure*}

\begin{figure*}[!t]
\setlength{\abovecaptionskip}{-0cm}
\setlength{\belowcaptionskip}{-.2cm}
\begin{center}
    \includegraphics[width=0.8\linewidth]{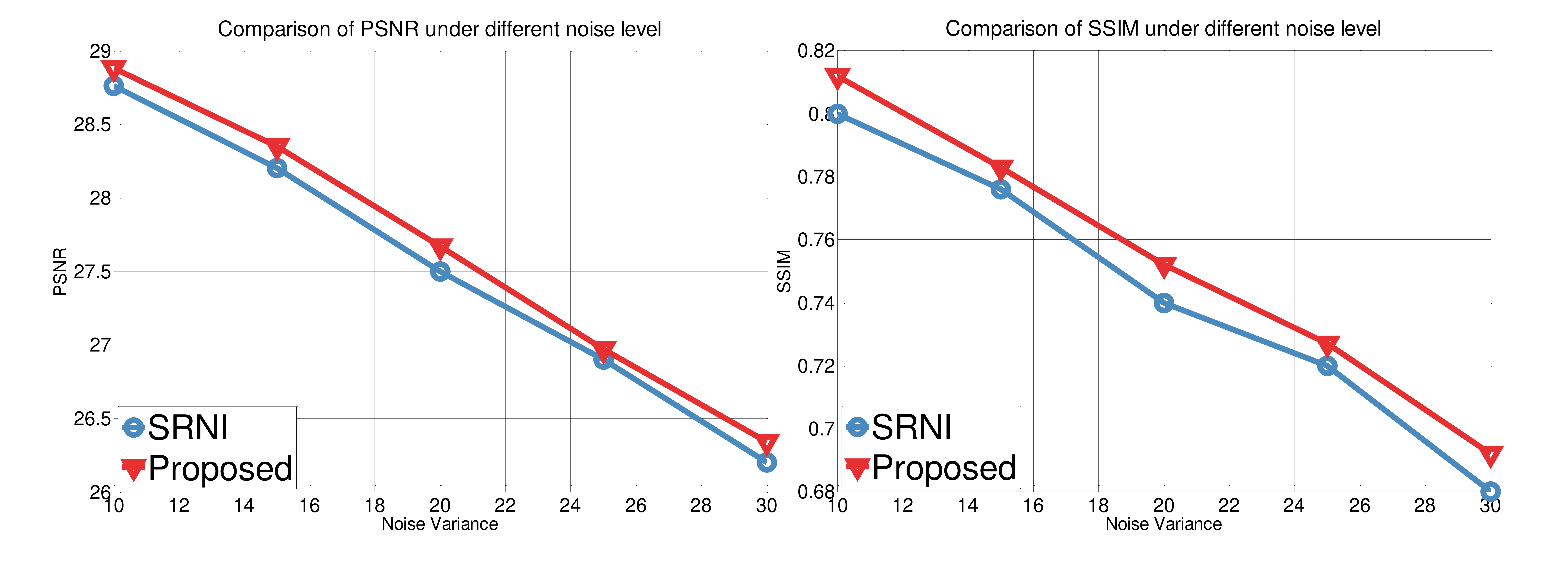}
\caption{PSNR and SSIM of SRNI and our proposed method on 50 noisy images from BSD300 under varied noise levels (from 10 up to 30).}
 \label{fig:compare_SRNI}
\end{center}
\end{figure*}


\subsection{The Experiment on Low-Rank Input Quantity vs. Its Performance}
\label{sec:5.2}
In the section \ref{set:4.3.2}, our theoretical proof shows that the relation between the low-rank input quantity and its performance should not be monotonically increasing. When input (the matrix $X$ stacked by the internal and external SR results) has very few images, adding more preliminary HR images will improve the overall performance until reaching the best, namely, the turning point. Afterward, more inputs may make performance decline. Unfortunately, it is nontrivial to theoretically prove how many input images could let the algorithm reach the turning point, because the errors and noise of each preliminary HR image varies according to the learning methods and data. To have an empirical result, we produce multiple preliminary HR images by setting $k \in \{1,\ldots, 8\}$ similar patches for the internal learning, $m\in \{1,\ldots, 8\}$ dictionary pairs for the external learning, and rotating patch at angles ($0^{\circ}, 90^{\circ}, 180^{\circ}, 270^{\circ}$). Then, the average performance of 27 test images against the number of input images are plotted in Fig. \ref{quantity_performance}. It should be noted that internal learning and external learning provide the same amount of samples as the low-rank input. This result verifies our proof in the section \ref{set:4.3.2}. One can see that the proposed solution improves the overall performance quickly to the maximum before the number of inputs reaches 36. It indicates that around 36 preliminary HR images\footnote{The number of inputs may vary with a small perturbation when applying different internal and/or external learning methods.} (including 18 external learning results and 18 internal ones) can result in the best output. This number ensures the genuine details appear more than $\mathcal{O}(r\log n)\approx$ 5 or 6 times in inputs. However, when more inputs are sustainedly added, the performance gradually decreases. This small required number of inputs makes it simple to tailor the internal and external learning methods without creating many preliminary HR images. We hereafter apply 36 HR images as the input in the following experiments.


\begin{figure*}[!t]
\setlength{\abovecaptionskip}{-0cm}
\setlength{\belowcaptionskip}{-.2cm}
\centering
    \includegraphics[width=.9\linewidth]{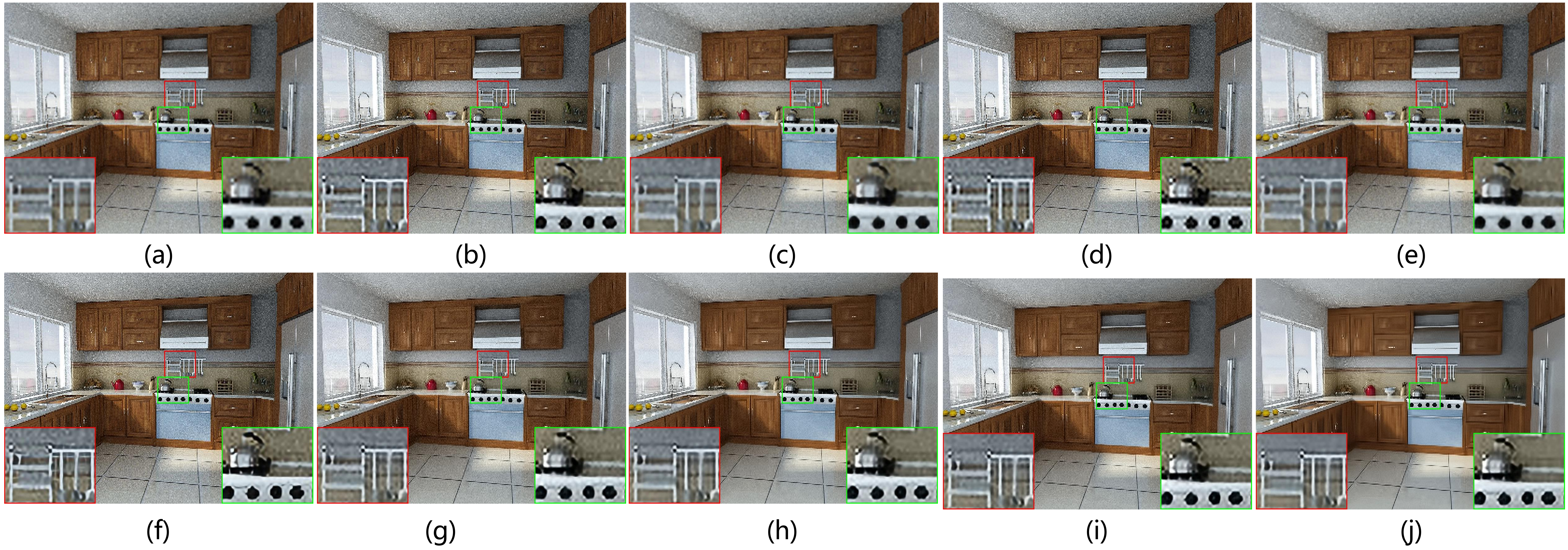}
\caption{Results of real noisy image (magnified by a factor of 3). Local magnifications are shown in the left bottom and right bottom of each example. (a) Bicubic interpolation. (b) A+ \cite{timofte2014a+}. (c) SRCNN \cite{dong2014learning}. (d) SCN \cite{liu2016robust}. (e) FSRCNN \cite{dong2016accelerating}. (f) VDSR \cite{Kim_2016_VDSR}. (g) TSE \cite{huang2015single}. (h) External$_\text{LRS}$. (i) Internal$_\text{LRS}$. (j) Proposed method.}\label{real_noise1}
\end{figure*}

\subsection{Comparison on Noiseless Images}
\label{set:5.3}
To demonstrate the improved performance, we compare the proposed method with A+ \cite{timofte2014a+}, TSE \cite{huang2015single}, SRCNN \cite{dong2014learning}, SCN \cite{liu2016robust}, FSRCNN \cite{dong2016accelerating}, VDSR \cite{Kim_2016_VDSR}, External$_\text{LRS}$, and Internal$_\text{LRS}$ on noiseless images from aforementioned three datasets. The quantitative comparison is listed in Table \ref{tab:table1}, from which we can see that our proposed method boosts PSNR and SSIM, outperforms the A+, SRCNN and TSE but performs worse than the deep learning based methods: SCN, FSRCNN and VDSR. Interestingly, only applying low-rank on the external learning improves PSNR 0.13 dB (SSIM 0.0017) than original method. Compared to others, it performs a little worse than SRCNN but better than A+, 0.10 dB on PSNR (0.0011 on SSIM). Similarly, applying low-rank on the internal learning improves PSNR 0.14 dB (SSIM 0.0016) than original one, and performs better than the best internal learning method, TSE. Furthermore, we apply low-rank on the integration of external and internal learning, and obtain better results, both in PSNR and SSIM. However, we have to admit that our proposed method are indeed worse than the deep learning based methods, particularly SCN and VDSR. Due to the conclusion, in Section \ref{sec:3} and \ref{sec:4}, external and internal learning methods produce different but complementary details. Our low-rank solution is able to wisely integrate pros of both methods and further boosts up SR performance. However, the SR result relies on the selected external and internal learning methods because low-rank itself does not produce new details. The major reason of the success of deep learning methods is that of having a more complete solution space by using huge training data and computation cost. Instead, the method in our solution searches less space and then results in a worse quantitative assessment in Table 1. Nevertheless, the more sophisticated the learning methods applied in our low-rank solution, the better results one will have. To have a better understanding, Fig. \ref{figure1} shows a qualitative comparison, for more results, please refer to Fig. 1 $\sim$ Fig. 2 in supplementary material. One can see that A+ produces some blur and jaggy artifacts, particularly, noticeable artifacts can be found along the edges. SRCNN produces HR image with relatively rich details compared with A+, but still creates evident artefacts along the edges. TSE applies an advanced patch matching strategy, thus achieves more visually pleasing results, where edges become sharper but artefacts mostly vanish. FSRCNN and VDSR generate more high-frequency details without noticeable jaggy artefacts. By applying low-rank on the external/internal learning methods individually, we successfully grasp the diversity of each preliminary HD image. Thus, the External$_\text{LRS}$ / Internal$_\text{LRS}$ achieves competitive visual results. One can see that External$_\text{LRS}$ has a better detail recovery when the patches frequency appear in the training datasets, and Internal$_\text{LRS}$ does perform better on the repeated patterns. Our proposed method, integrating the pros of the external and internal learning methods, generates a more visually pleasing image. Although our method is worse than SCN, FSRCNN and VDSR in quantitative comparison, the qualitative  results are rather comparable.


\begin{figure*}[!t]
\centering
    \includegraphics[width=.9\linewidth]{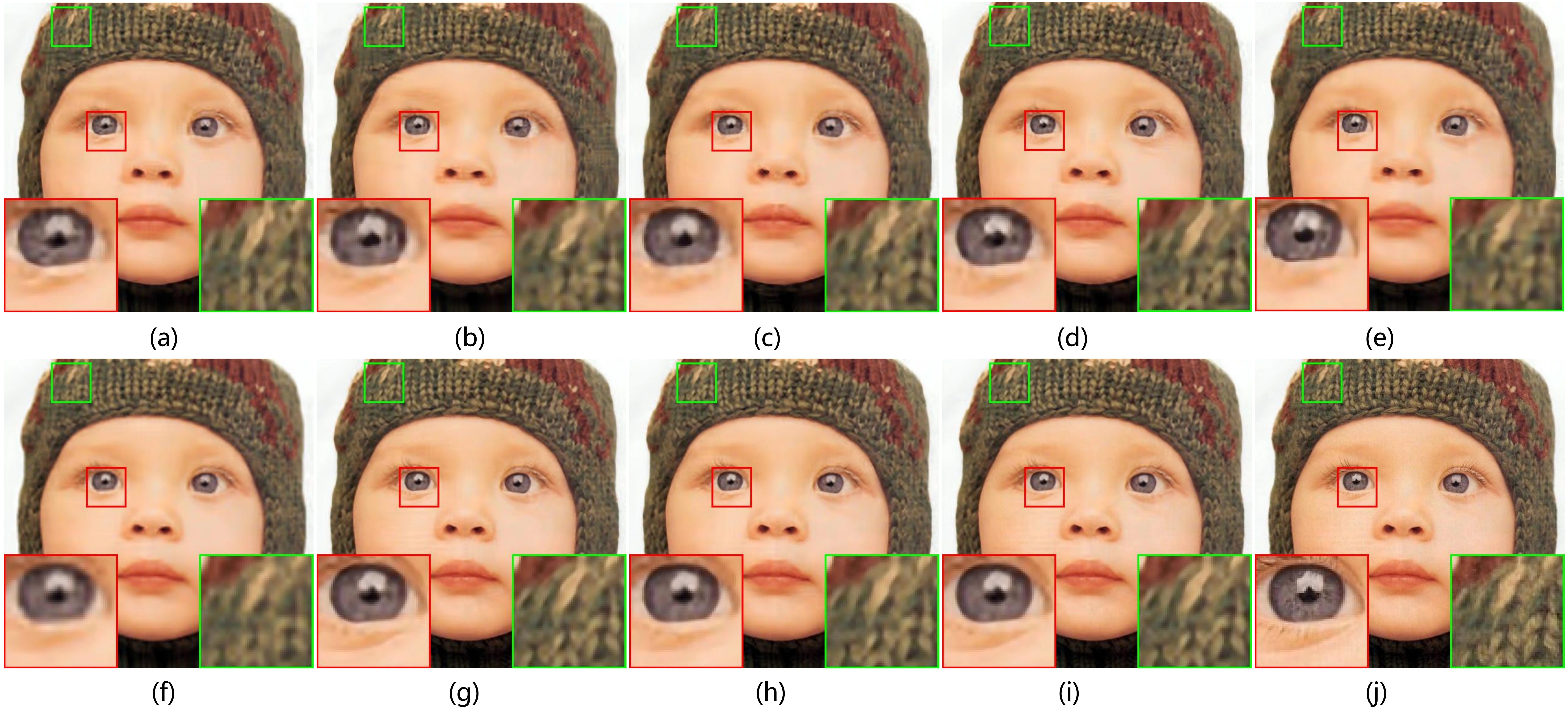}\vspace{-.3cm}
\caption{Results of synthetic noisy image, \emph{baby}, with standard deviation of 8 (magnified by a factor of 3). Local magnifications are shown in the left bottom and right top of each example. (a) BM3D + A+ \cite{timofte2014a+}. (b) BM3D + SRCNN \cite{dong2014learning}. (c) BM3D + SCN \cite{liu2016robust}. (d) BM3D + FSRCNN \cite{dong2016accelerating}. (e) BM3D + VDSR \cite{Kim_2016_VDSR}. (f) BM3D + TSE \cite{huang2015single}. (g) External$_\text{LRS}$. (h) Internal$_\text{LRS}$. (i) Proposed method. (j) Ground truth.}\label{denoised_8}
\end{figure*}

\begin{figure*}[!t]
\setlength{\abovecaptionskip}{0.cm}
\setlength{\belowcaptionskip}{-.7cm}
\begin{center}
    \includegraphics[width=.9\linewidth]{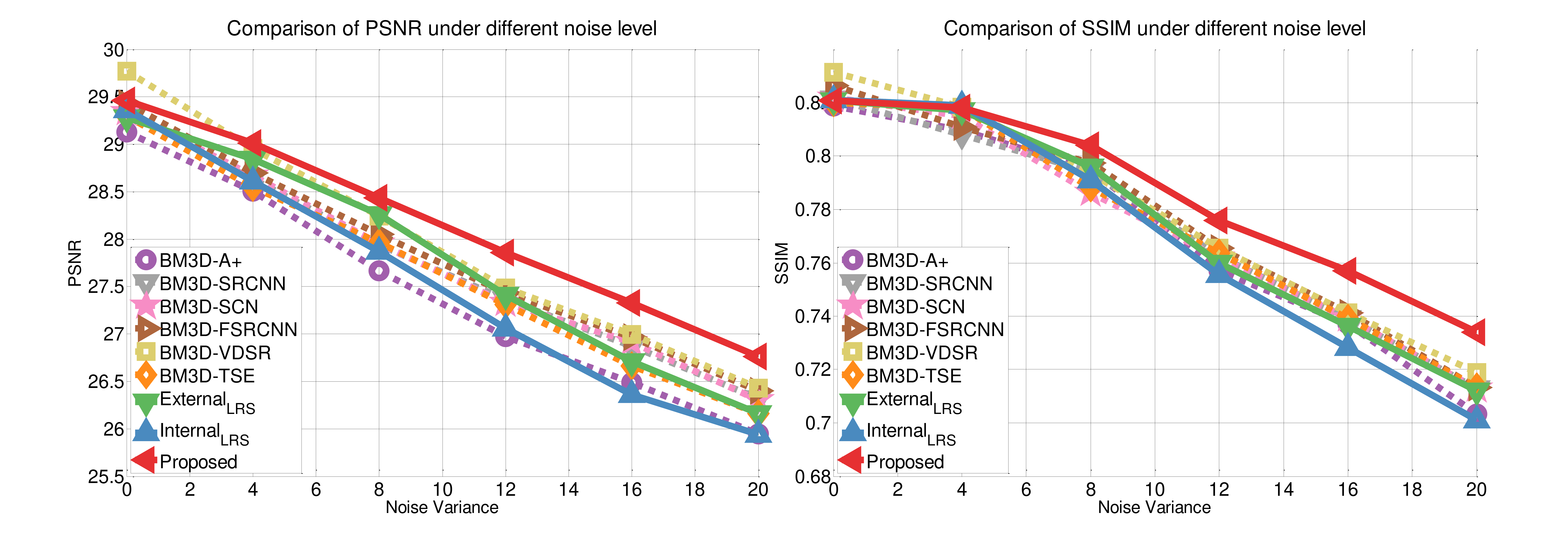}\vspace{-.3cm}
\caption{PSNR and SSIM of all comparative methods under varied noise levels. The noisy images are denoised by BM3D first, then super-resolved by six competing methods. Instead, External$_\text{LRS}$, Internal$_\text{LRS}$ and our proposed method do not take the denoising process.}
  \label{fig:denoised_psnr_ssim}
\end{center}
\end{figure*}

\subsection{Comparisons on the Noisy Images}
\label{set:5.4}

In reality, the LR images are often corrupted by noise. The analysis in section \ref{set:4.3.2} shows that our low-rank solution has already taken the noise into account. The experiments below, probe the lowest nature of the low-rank solution to a variety of noisy data in SR. 

\subsubsection{The synthetic noisy images}
\label{sec:5.4.1}
We take 14 images from Set14, and then add Gaussian noise with the variance from 4 to 20 where the step is 4. Comparison is done among A+ \cite{timofte2014a+}, SRCNN \cite{dong2014learning}, SCN \cite{liu2016robust}, FSRCNN \cite{dong2016accelerating}, VDSR \cite{Kim_2016_VDSR}, TSE \cite{huang2015single}, External$_\text{LRS}$, Internal$_\text{LRS}$, and our proposed method, where SCN claims its robustness to noise. Fig. \ref{fig:noise_psnr_ssim} plots the PSNR and SSIM scores against varied noise levels, and shows the quality of all images decreases with the noise level increasing. But a significant difference is that the proposed method decreases much slower than the other six. This result proves that the low rank solution can effectively remove noise.

The qualitative results are shown in Fig. \ref{noise_20} and Fig. \ref{noise_12}, where the LR \emph{face} and \emph{monarch} images are added Gaussian white noise with standard deviation of 20 and 12 respectively, for more results, please refer to Fig. 3 $\sim$ Fig. 5 in supplementary material. One can see that SRCNN, FSRCNN and VDSR are sensitive to severe noise. A+ method successfully suppresses the noise, but more artefacts appear in smooth regions. SCN method performs slightly better than above methods, but remaining noise is still obvious. TSE method produces a relatively finer result by patch-matching strategy. Our proposed method suppresses the noise significantly and produces little noise-caused artifacts. It could be explained that all competing methods do not discriminate the noise, but process it as the real detail during SR procedure. On the contrary, Equation 8 shows that low-rank decomposition is able to separate the sparse component (noise and error) from inputs.

Furthermore, we compare with SRNI \cite{singh2014super} that is devoted to super-resolve noisy image. However, we sampled 50 images from BSD300 but excluded from BSD100 which is one of benchmark databases. For a fair comparison, we follow the exact same experimental setting in \cite{singh2014super} by downsampling those images, adding Gaussian noise with the variance from 10 up to 30, and then super-resolving the noisy LR images. Fig. \ref{fig:compare_SRNI} lists the PSNR and SSIM scores against varied noise levels, and shows our proposed method consistently outperforms SRNI across all testing noise levels.

\subsubsection{The real noisy images}
\label{sec:5.4.2}
We find the noise in real LR images may not be always Gaussian, thus choose the real noisy images from \cite{neatvideo} for a further comparison. Since there is no ground truth, we are not able to provide the quantitative comparison. The qualitative comparison in Fig. \ref{real_noise1} is similar as the results of the synthetic noisy images. For more results, please refer to Fig. 6 $\sim$ Fig. 8 in supplementary material. One can see the competing methods are still sensitive to severe noise in real noisy images.  SRCNN produces some blur along the edges but VDSR generates unrealistic sharp details. Instead, our proposed method consistently suppresses the noise and recover the edges better than the others.



\subsubsection{The denoised noisy images}
\label{sec:5.4.3}
When encountering noisy LR images, a straightforward consideration is to denoise them and then apply SR. To test this configuration, we compare our proposed method, External$_\text{LRS}$ and Internal$_\text{LRS}$ with denoising + SR. The SR methods and testing data are the same as those in Section \ref{sec:5.4.1}. The denoising process is done by the well accepted algorithm BM3D \cite{dabov2007image}. Fig. \ref{fig:denoised_psnr_ssim} plots the PSNR and SSIM scores against varied noise levels, and illustrates the denoising process does improve the competing methods on noisy LR images. This configuration even achieves comparable performance as External$_\text{LRS}$ and Internal$_\text{LRS}$. However, when applying low-rank on both external and internal learning methods, our proposed method is superior to them again due to an effective integration of the complementary details recovered by two methods. Fig. \ref{denoised_8} shows the qualitative results of a test image with noise level 8, for more results, please refer to Fig. 9 $\sim$ Fig. 11 in supplementary material. Thanks to BM3D denoising, the majority of noise has vanished from results of all competing SR methods. However, the local magnifications depict certain high-frequency signal along edges is inevitably lost because of denoising procedure, and certain residual noise and artifacts are magnified. Instead, our proposed method has no denoising process. Thus, it can not only suppress the noise, but also preserves the high-frequency details.

\section{Conclusion}
\label{sec:6}
Instead of solving the problem of SR by internal learning or external learning separately, we proposed a low-rank solution to integrate their pros together. We show that the attributes of internal and external learning are complementary in the feature space and image plane. Meanwhile, their estimation errors are sparse. This is the basis, upon which the proposed solution is found. With theoretical analysis and a real data experiment, we also proved that the low-rank solution does not require massive input to achieve a desired SR image. This result makes tailoring the internal and external learning methods easier for the integration. Compared to other state-of-the-art methods, our proposed solution is parameter free, and does not  need preprocessing for internal or external prior selection during the integration. The experiment on the noiseless data has demonstrated a comparable performance of our proposed solution. Particularly, on a variety of noisy data, our solution performs superior on restraining noise and recovering sharp details. Furthermore, additional performance could be gained by generalizing our solution with more recent methods. Till now, it is still uncertain if the numbers of preliminary HD images from internal and external learning methods must be same. This is an issue we hope to further explore in future studies in order to determines better performance.

%
%
%
%

\small
\bibliographystyle{IEEEtran} \vspace{-4pt}
\bibliography{referencesr}
\end{document}